\documentclass[runningheads]{llncs}
\usepackage{graphicx}
\usepackage{xcolor}
\usepackage[normalem]{ulem} 
\usepackage{cite}
\usepackage{hyperref}
\usepackage{color}
\usepackage[ruled,vlined]{algorithm2e}
\usepackage{bm}
\usepackage{amsmath}
\usepackage{amssymb}
\usepackage{caption}
\usepackage{subcaption}
\usepackage{wrapfig}
\usepackage{soul}
\usepackage[inline]{enumitem}
\usepackage{float}
\usepackage[draft]{todonotes} 

\usepackage{tabularx}
\usepackage{pifont}
\newcommand{\cmark}{\ding{51}}%
\newcommand{\xmark}{\ding{55}}%

\definecolor{darkgreen}{rgb}{0.0, 0.42, 0.24}

\newcommand{\accval}{\textit{acc}\textsubscript{\textit{val}}}
\newcommand{\acctest}{\textit{acc}\textsubscript{\textit{test}}}

\begin{document}

\title{Local Search is a Remarkably Strong Baseline for Neural Architecture Search
}
\titlerunning{Local Search is a Remarkably Strong Baseline for Neural Architecture Search}
%

\author{T. Den Ottelander\inst{1} \and
A. Dushatskiy\inst{1},
M. Virgolin\inst{1} \and
P.A.N. Bosman\inst{1,2}}
\authorrunning{T. Den Ottelander et al.}
%
\institute{Centrum Wiskunde \& Informatica, Amsterdam, the Netherlands\\
\email{\{tdo,arkadiy.dushatskiy,marco.virgolin,peter.bosman\}@cwi.nl}
\and
Delft University of Technology, Delft, the Netherlands
}
\maketitle              
\begin{abstract}
Neural Architecture Search (NAS), i.e., the automation of neural network design, has gained much popularity in recent years with increasingly complex search algorithms being proposed. Yet, solid comparisons with simple baselines are often missing. At the same time, recent retrospective studies have found many new algorithms to be no better than random search (RS). In this work we consider, for the first time, a simple Local Search (LS) algorithm for NAS. We particularly consider a multi-objective NAS formulation, with network accuracy and network complexity as two objectives, as understanding the trade-off between these two objectives is arguably the most interesting aspect of NAS. The proposed LS algorithm is compared with RS and two evolutionary algorithms (EAs), as these are often heralded as being ideal for multi-objective optimization. To promote reproducibility, we create and release two benchmark datasets, named MacroNAS-C10 and MacroNAS-C100, containing 200K saved network evaluations for two established image classification tasks, CIFAR-10 and CIFAR-100. Our benchmarks are designed to be complementary to existing benchmarks, especially in that they are better suited for multi-objective search. We additionally consider a version of the problem with a much larger architecture space. While we find and show that the considered algorithms explore the search space in fundamentally different ways, we also find that LS substantially outperforms RS and even performs nearly as good as state-of-the-art EAs. We believe that this provides strong evidence that LS is truly a competitive baseline for NAS against which new NAS algorithms should be benchmarked.

\end{abstract}
\section{Introduction}
    
Deep learning has achieved excellent results for machine learning tasks in heterogeneous fields, including image recognition~\cite{russakovsky2015imagenet}, natural language processing~\cite{radford2019language}, games~\cite{silver2017masteringGo,alphastarblog}, and genomics~\cite{senior2020deepmind_protein}. While the availability of large amounts of data~\cite{kitchin2014data} and hardware advancements (e.g., tensor parallel processing units~\cite{jouppi2017datacenter}) have been key enabling factors for deep learning, the success of the field is also due to the ingenious design of competent neural network architectures (consider, e.g., the invention of residual connections~\cite{he2016deep}).
    
Proper and innovative architecture design requires experience, intuition, and expensive trial-and-error. This has sparked research on techniques to automate this task, i.e., the field of Neural Architecture Search (NAS) has emerged~\cite{Elsken2018NASSurvey,wistuba2019nassurvey}. 
NAS research is quickly gaining popularity: 2019 alone counts almost 250 publications\footnote{\url{https://www.automl.org/automl/literature-on-neural-architecture-search/}}. 
Most NAS proposals present new, typically increasingly complex, NAS algorithms. However, recent work questions whether the need for many of these algorithms is actually justified:  \cite{Yu2020Evaluating} and \cite{Yang2020NASishard} showed that several NAS algorithms are no better than Random Search (RS) when validated properly. 

While RS can be considered perhaps the simplest form of a baseline, it is also by far the worst form of heuristic search in many cases, with the worst scalability. For this reason, it is often not even considered in many modern heuristic design studies. Yet, from RS to more complex search heuristics such as EAs, is still quite a leap. In the gap for instance lie almost equally simple and classical, yet far less random in nature, Local Search (LS) techniques. Still LS has, to the best of our knowledge, not been applied to NAS before, except for in~\cite{white2020local}, a work that happened independently and concurrently to ours (see Sec.~\ref{sec:related}). 

In this paper, we consider LS for NAS and in doing so, we make three contributions:
\begin{enumerate*}[label=(\arabic*)]
    \item We propose a simple and parameter-less LS algorithm for multi-objective (accuracy vs. architecture complexity) NAS.
    \item To enable quick and easy benchmarking, we release\footnote{ Benchmark datasets: \url{https://github.com/ArkadiyD/MacroNASBenchmark} \\ Source code for reproducibility: \url{https://github.com/tdenottelander/MacroNAS}} two datasets containing cached performances for over 200,000 Convolutional Neural Network (CNN) architectures for CIFAR-10 and CIFAR-100. Our datasets, named MacroNAS-C10 and MacroNAS-C100, are designed to be complementary to similar existing proposals \cite{Ying2019NAS-Bench-101,dong2020nasbench201}. 
    We remark that we designed these datasets to be able to compare NAS approaches, and not necessarily to obtain State-of-the-Art (SotA) neural networks.
    \item We provide evidence on the validity of LS being a superior baseline for NAS by conducting several experiments. In particular, we consider NAS in different search spaces; We compare LS with RS as well as with a classic and a SotA Evolutionary Algorithm (EA); We include an experiment on CIFAR-100 where the possible architectures are less constrained, leading to a large search space containing over 104 Billion possible neural architectures.
\end{enumerate*}

In this work, we consider NAS in a multi-objective setting. Single-objective NAS, i.e., searching only for good accuracy-networks, has already been amply investigated in \cite{Yu2020Evaluating,Yang2020NASishard}, where it was found that very different search algorithms, including RS, actually perform very similar.
Nevertheless, additional experiments that show the performance of LS for single-objective NAS are included in the Appendix and provide similar insights to those found in~\cite{white2020local}.
Multi-objective NAS on the other hand has been studied less (see Sec.~\ref{sec:related}), and is arguably a more interesting setting, where sophisticated search algorithms may potentially be needed. Alongside RS, we consider EAs, since they are a natural fit for multi-objective optimization, and are often used for NAS.

\section{Related work} \label{sec:related}


NAS can be broadly categorized by the level at which the search is performed, i.e., macro-level and micro-level~\cite{Elsken2018NASSurvey} (different nomenclatures exist~\cite{wistuba2019nassurvey}). Macro-level search aims at composing (wiring) atomic \emph{cells} of different type. A cell is a computational graph (a sub-net) composed of different operations/layers (convolutions, activation functions, etc.). In micro-level search, the position of cell types within the architecture is pre-decided (often a same cell type is repeated in multiple places), while it is the cells' internal wiring that is optimized. Both levels can also be searched at the same time~\cite{Lu2018NSGA-Net, miikkulainen2019codeepneat}, potentially along with hyper-parameter settings~\cite{wong2018transfer, miikkulainen2019codeepneat}. In the remainder of the paper, we will refer to ``architecture'' and ``network'' (for brevity, ``net'') interchangeably.


Search algorithms considered for NAS vary wildly. Reinforcement Learning (RL), Bayesian optimization approaches, and EAs are among the most popular~\cite{Elsken2018NASSurvey}; the latter being typically employed in multi-objective scenarios.
In this paper, we mostly focus on performing multi-objective NAS at macro-level. Therefore, in the following we report on recent works in this direction, and focus on whether comparisons against a baseline (i.e. RS) were made. 

The work in both~\cite{Kim2017NEMO,Chu2019MOREMNAS} builds upon NSGA-II~\cite{Deb2002NSGA-II} to obtain a front of nets that trade off accuracy and compactness. Although nets with SotA performance are ultimately obtained, the search algorithm is not compared with any baseline.
In~\cite{Elsken2018LEMONADE}, a multi-objective EA is proposed that includes Lamarckian mechanisms: instead of training nets from scratch, child nets inherit the weights of their parents and are only fine-tuned.
Although the method is shown to outperform RS, it is unclear whether this is due to a difference with the way the EA searches, or solely because of weight inheritance.
In~\cite{Lu2018NSGA-Net}, NSGA-Net is presented, which evolves fixed-length architectures at both macro- and micro-level. The authors show, through experimental analysis, that NSGA-Net is superior to RS.
Summarizing, most of the literature work considers only RS as a baseline (sometimes in not entirely equal conditions), or no baseline at all. 

Importantly, the only work we are aware of that investigated the use of an LS algorithm is~\cite{white2020local}, which appeared on arXiv shortly after the first version of our paper appeared on the same platform. The work shares strong similarities with ours, as it investigates the use of a simple LS algorithm for NAS, and reaches the same conclusion, i.e., that LS is a strong baseline for NAS. Despite these similarities, our works can be considered complementary to each other: the authors of~\cite{white2020local} focus on single-objective NAS benchmarks and provide interesting theoretical underpinnings; we instead focus on multi-objective NAS, and, among other contributions, propose new NAS benchmark datasets.


    

Since we release benchmark datasets of saved net evaluations for macro-level NAS, we also refer to important related works of this nature. NAS-Bench-101~\cite{Ying2019NAS-Bench-101} contains cached performance indicators for 423,000 nets on CIFAR-10, while NAS-Bench-201~\cite{dong2020nasbench201} does the same also for CIFAR-100 and ImageNet, but for far less architectures (15,625). Both datasets are built for micro-level NAS. We did not consider NAS-Bench-201 because the limited number of possible architectures can limit the analysis of the search behavior for long run-times. We built new benchmark datasets for three reasons. First, we intended to provide the community with a macro-level search benchmark dataset for better reproducible NAS: we are not aware of other works proposing this. Second, NAS-Bench-101 includes infeasible solutions (the MacroNAS datasets do not), and choosing how to handle them substantially changes the search behavior of an algorithm. Lastly, for NAS-Bench-101 it has already been shown that RS is a competitive baseline in a single-objective setting~\cite{Ying2019NAS-Bench-101}. We will show however that, in fact, in the search space of NAS-Bench-101 it is hard to do any better than RS, even when considering a multi-objective setting (see Sec.~\ref{sec:exp-preliminary}). 

\section{Objectives, Encodings, Search algorithms}\label{sec:obj-enc-algs}
We now present the objectives to optimize, the encoding schemes we use, and the algorithms we consider in the experimental analysis. For a detailed description of the MacroNAS datasets we introduce, we refer the reader to the Appendix.

\subsection{Objectives}
The first objective $f_1$ is to maximize accuracy (fraction of correct net predictions). Particularly, we set $f_1$ to the validation accuracy (\accval), i.e., the accuracy of the net for a set of data that is not used for training the net, to assess whether the net generalizes. Because the validation set is still used in the NAS optimization loop, we consider the ability of nets with good \accval{} to generalize to a second set of completely held-out data, i.e., the test set, in Sec.~\ref{sec:generalization}.

The second objective $f_2$ is to maximize a net's efficiency, i.e., to minimize the Mega Multiply--ACcumulate operations (MMACs, 1 MMAC $\simeq $ 2 FLOPs), which is the number of GPU computations a net requires to make a prediction. MMACs is a popular complexity metric~\cite{sze2017efficient} and is deterministic (in contrast to actual time taken). Because large differences in scale of the objectives can influence search mechanisms (e.g., the crowding distance~\cite{Deb2002NSGA-II}), we set MMACs to be in the same range as \accval{}, i.e., we normalize MMACs to lie in $[0,1]$. The min/max MMAC values for the normalization are trivial to determine as it suffices to consider the smallest/largest achievable nets. 
For the sake of a consistent optimization direction in both objectives, the secondary objective $f_2$ becomes maximizing $1 - \textit{normalize}(\text{MMACs})$.
Since no MMACs were reported for NAS-Bench-101, for the experiment in Sec.~\ref{sec:exp-preliminary} we use $1 - \textit{normalize}(\textit{\#parameters})$.


\subsection{Encodings}

We use a direct encoding that represents feed-forward CNNs. The encoding is an array of discrete variables $[ x_1$, \dots, $x_\ell]$, where each $x_i$ takes values (cell types) in $\Omega_i$, a position-dependent alphabet.


\begin{figure}[ht]
    \vspace{-0.4cm}
    \centering
    \begin{subfigure}{.73\textwidth}
      \centering
      \includegraphics[width=\linewidth]{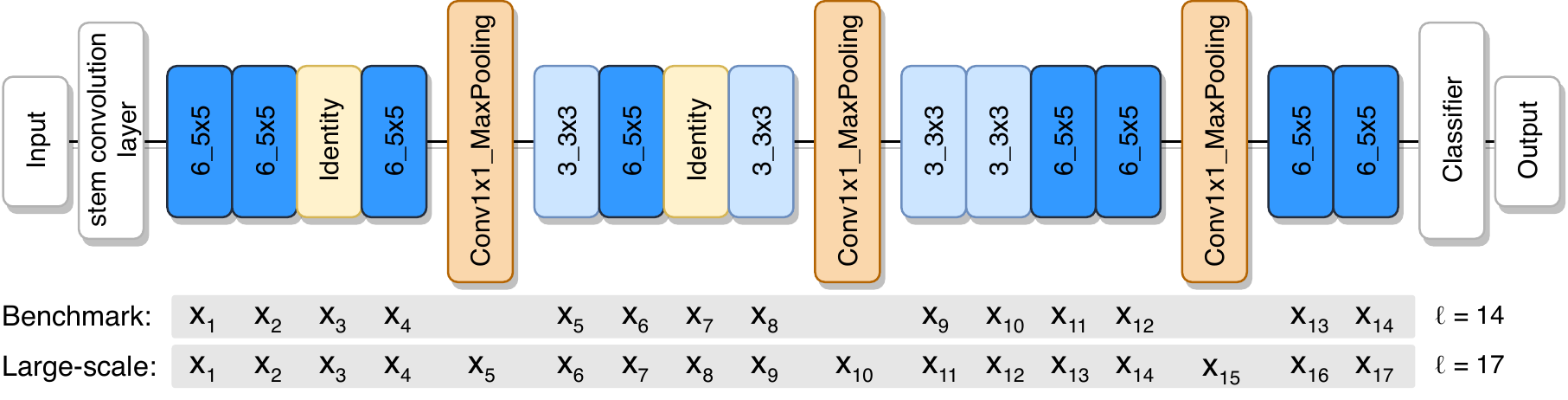}
    \end{subfigure}%
    \hspace{5pt}
    \begin{subfigure}{.24\textwidth}
      \centering
      \includegraphics[width=\linewidth]{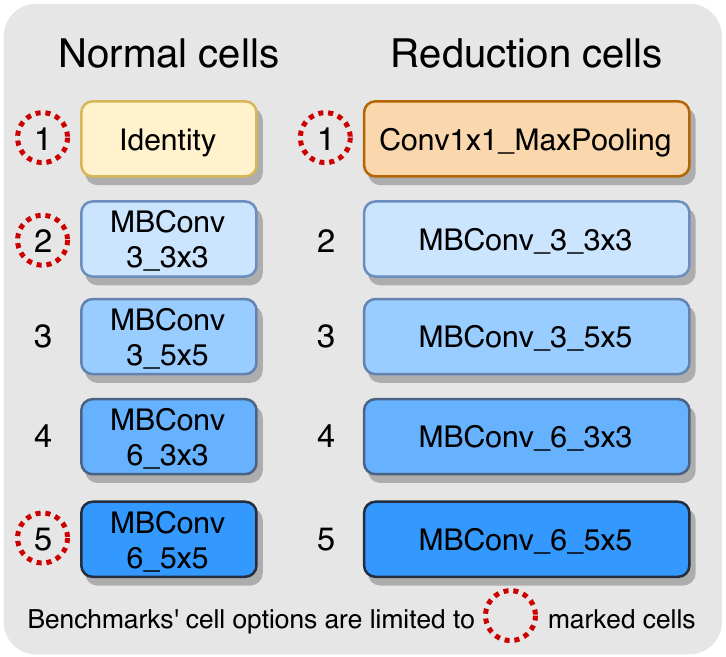}
    \end{subfigure}
    \vspace{-.3cm}
    \caption{Best \accval{} net on CIFAR-10 (left) and cell types (right). For MacroNAS, 3 types are possible for normal cells (types 1, 2 and 5) and reduction cells (positions 5, 10, and 15) are fixed to type 1. For the large-scale experiment, 5 types are available for all cells. 
    For details on cell types see the Appendix.}
    \vspace{-0.6cm}
    \label{fig:example-search-space}
\end{figure}


In the MacroNAS datasets, the number of searchable cells (length of the encoding) is $\ell=14$. In particular, only \emph{normal} cells, i.e., cells that maintain the height and width of the input they receive \cite{zoph2018learning}, can be optimized. Conversely, cells that reduce spatial dimensions (\emph{reduction} cells) are fixed. 
The alphabet $\Omega_i$ is identical for each position $i$ and contains 3 options. One option is the \emph{identity}, i.e., a placeholder cell that forwards information as is. The other two options are commonly used convolutional blocks from literature (described in the Appendix). Fig.~\ref{fig:example-search-space} shows an example of an encoding.
Even though the encoding has a fixed length, it can represent variable-length architectures by means of identity cells. Note that the mapping from encoding to (computationally unique) architecture is redundant. The size of the search space (total number of encodings) is $3^{14}$ = 4,782,969, which maps to 208,537 architectures. The latter number is sufficiently small to allow us to evaluate all architectures and save their performances to create the benchmark datasets.

For the large-scale experiment, we increase the number of variables $\ell$ to 17 by including the optimization of reduction cells, and also increasing the alphabet size $|\Omega_i|$ to 5, $\forall i$. Here, the position $i$ influences what $\Omega_i$ is used, as options for normal cells are different from options for reduction cells (see Fig.~\ref{fig:example-search-space}). The total search space of $5^{17}$ = 762,939,453,125 maps to 104,086,030,125 architectures.

The encoding for NAS-Bench-101 \cite{Ying2019NAS-Bench-101} (micro-level) is different. Here, both the type of operations in a (repeated) cell, and their connections, can be optimized. The encoding is a binary string of length $21$ that represents connections, concatenated with a ternary string of length $5$ that represents operations.



\subsection{Search algorithms}\label{sec:search-algs}


The first algorithm we consider is NSGA-II~\cite{Deb2002NSGA-II}. It is by far the most popular multi-objective EA in general.
We use most commonly adopted parameter settings of NSGA-II: 2-point crossover, single-variable mutation with probability $p_m=1/\ell$, tournament size $2$ and population size $n_\text{pop} = 100$.

Fundamentally better algorithmic design is foundational to EA research. This means building better problem-specific EAs or ones capable of doing better at large classes of interesting problems. 
In the latter direction lie linkage learning EAs that attempt to automatically detect and exploit building blocks of non-trivial sizes~\cite{Thierens2011OptMixing,Pelikan2006Scalable}. To the best of our knowledge, such algorithms have not yet been tried for NAS. Studying if they have merit here as well is interesting.
Thus, we consider the Multi-Objective Gene-pool Optimal Mixing Evolutionary Algorithm (MO-GOMEA). MO-GOMEA mostly differs from a classic genetic algorithm in that every generation it computes a model of \emph{linkage}, i.e. the estimated strength of interdependency between variables, and uses this model to propagate potential building blocks during variation \cite{Thierens2011OptMixing}. 
If the NAS search space exhibits any linkage, this algorithm can likely exploit it. Also, MO-GOMEA uses clusters to partition the population in objective space, so it is generally able to improve upon specific parts of the front by recombining within niches.
We use one of the latest implementations for discrete optimization~\cite{Luong2018MO-GOMEA}, which includes the Interleaved Multi-start Scheme (IMS).
The IMS allows to avoid the manual setting of population size parameter by evolving populations of increasing sizes in interleaved fashion. For the initial population size, we use $n_\text{pop}=8$ (default).

\begin{wrapfigure}{r}{.3\linewidth}
    \vspace{-0.6cm}
    \centering
    \begin{subfigure}{\linewidth}
        \centering
        \includegraphics[width=\linewidth]{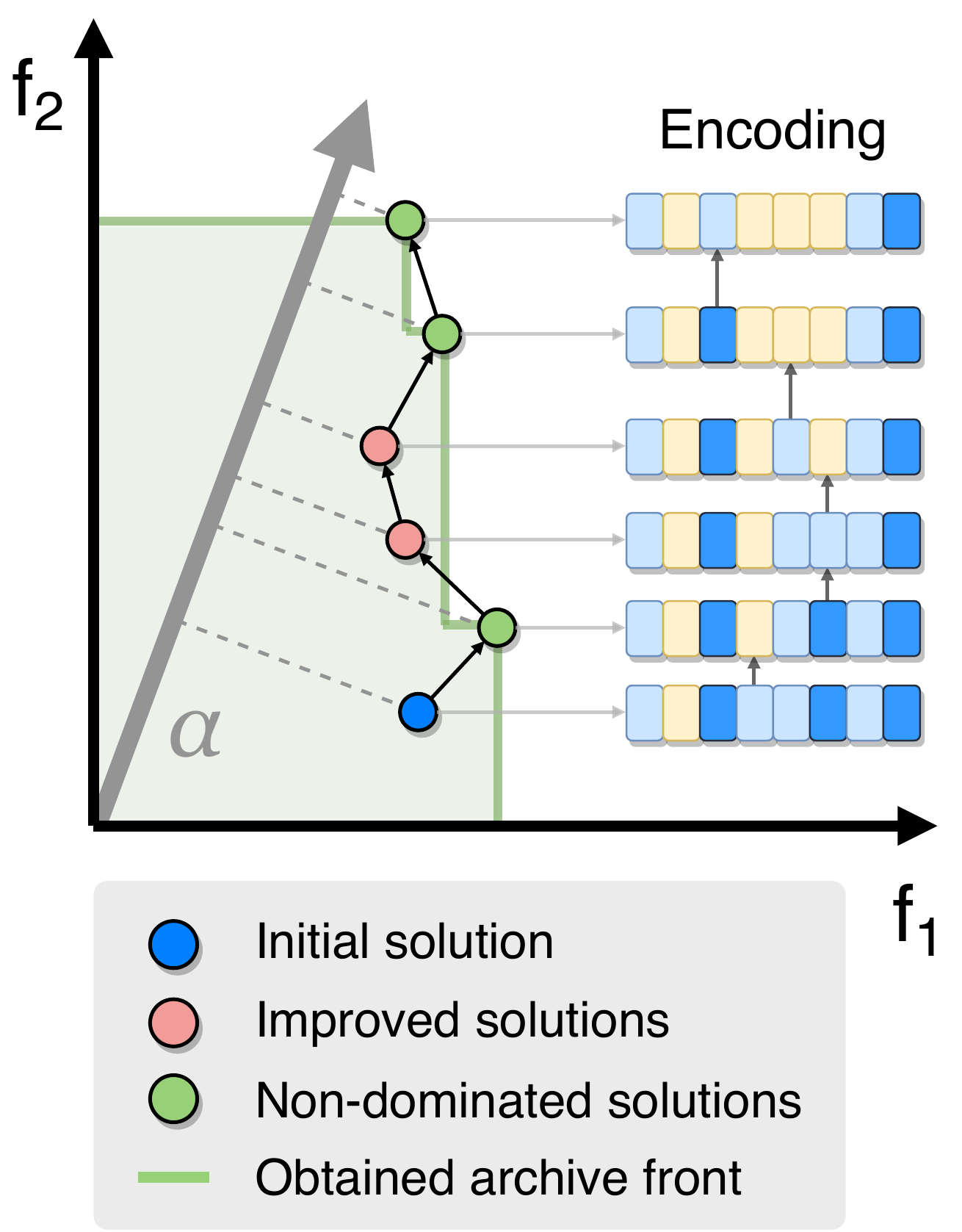}
    \end{subfigure}
    \vspace{-0.1cm}
    \caption{Example iteration on one random scalarization coefficient $\alpha$ in LS. Only solutions that improve the scalarized objective are shown.}
    \label{fig:LS}
    \vspace{-1cm}
\end{wrapfigure}

We further consider RS as a standard baseline. RS generates new nets repeatedly by sampling each cell uniformly at random from the alphabets $\Omega_i$.


Finally, we consider a simple random 
restart LS algorithm:
\vspace{-0.1cm}
\begin{enumerate}
    \item \label{list:LS-step1} Initialize a random architecture (as in RS);
    \item Sample a scalarization coefficient $\alpha \sim \mathcal{U}(0,1)$;
    \item Consider all variables in random order, as follows:
    \begin{enumerate}[label*=\arabic*.]
    \item For each variable $x_i$, evaluate the net obtained by setting $x_i$ to each option in $\Omega_i$. Keep the best according to $\alpha \times f_1 + (1 - \alpha) \times f_2$;
    \end{enumerate}
    \item Repeat step~\ref{list:LS-step1} until the computation budget (evaluations / time) is exhausted.
\end{enumerate}
\vspace{-0.1cm}

In other words, we perform one round of first-improvement LS at the level of variables, but best-improvement LS at the level of options for one variable. 
Fig.~\ref{fig:LS} illustrates one iteration of our LS algorithm (from now on, we simply refer to it as LS).

\section{Experimental setup}\label{sec:exp-setup}

All search algorithms store an archive $\mathcal{A}$ of best-found nets according to (strict) Pareto domination. Formally, it is:
$\bm{x} \succ \bm{y}$ (``$\bm{x}$ dominates $\bm{y}$'') iff $f_i(\bm{x}) \geq f_i (\bm{y}) \land \bm{f}(\bm{x}) \neq \bm{f}(\bm{y})$, for each objective~$f_i$.
The archive thus satisfies $\nexists \bm{x}, \bm{y} \in \mathcal{A}: \bm{x} \succ \bm{y}$. We use no size limitation for the archive, and update the archive each time a new net is discovered (the time cost is insignificant compared to evaluating a net's performance, see below).

To compare algorithms, we consider the improvement of the hypervolume enclosed by $\mathcal{A}$, using the origin as reference point. For the experiments of Secs.~\ref{sec:exp-preliminary} and~\ref{sec:exp-benchmarks-mo} (where the Pareto front is known), we also investigated the inverse generational distance~\cite{Bosman2010DistanceMetric}, and obtained very similar results (omitted for brevity). 
We only count the evaluations of unique architectures as evaluating a net is typically costly (minutes to hours). Caching is used to not re-evaluate a same architecture twice.

We handle the net with \emph{identity} cells for all variables differently from the others. This trivial net can be considered fundamentally uninteresting to search for, but it still influences multi-objective metrics.
We therefore set NSGA-II and MO-GOMEA to include one instance of it in the population at initialization. 
Similarly, we include it in the archive of RS and LS from the beginning.

For experiments concerning the benchmark datasets, i.e., where performances for different architectures are pre-stored, we perform 30 runs for each algorithm.
For the large-scale search space, where pre-evaluating all nets is simply unfeasible and a single evaluation takes on average 6.5 minutes on Nvidia RTX 2080 Ti GPU, we perform 6 runs for each algorithm.
To assess whether differences between algorithms are significant, we adopt pairwise non-parametric Mann-Whitney-U tests with Bonferroni correction.


\section{Results}\label{sec:results}
First, we compare the algorithms on NAS-Bench-101 and on MacroNAS-C10/ C100. Next, we analyze the performance on MacroNAS-C100, and on the large-scale version for this task, in more detail. Lastly, we discuss what the observed differences mean when considering the nets' ability to generalize.

\subsection{Preliminary experiments}\label{sec:exp-preliminary}

\begin{figure}[b]
    \vspace{-0.4cm}
    \centering
    \begin{subfigure}{0.32\linewidth}
        \centering
         \includegraphics[width=\linewidth]{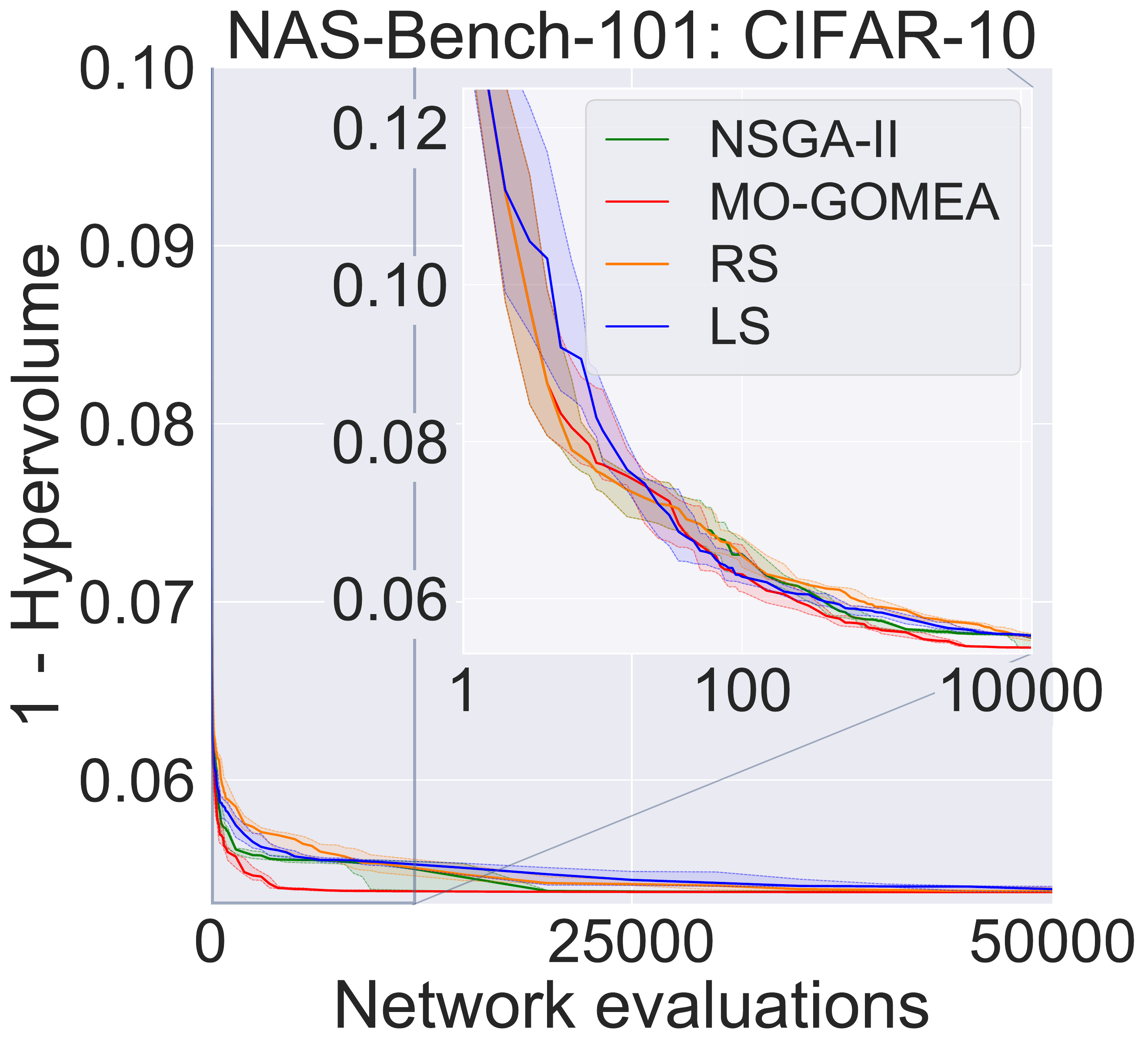}
    \end{subfigure}
    \begin{subfigure}{0.32\linewidth}
        \centering
         \includegraphics[width=\linewidth]{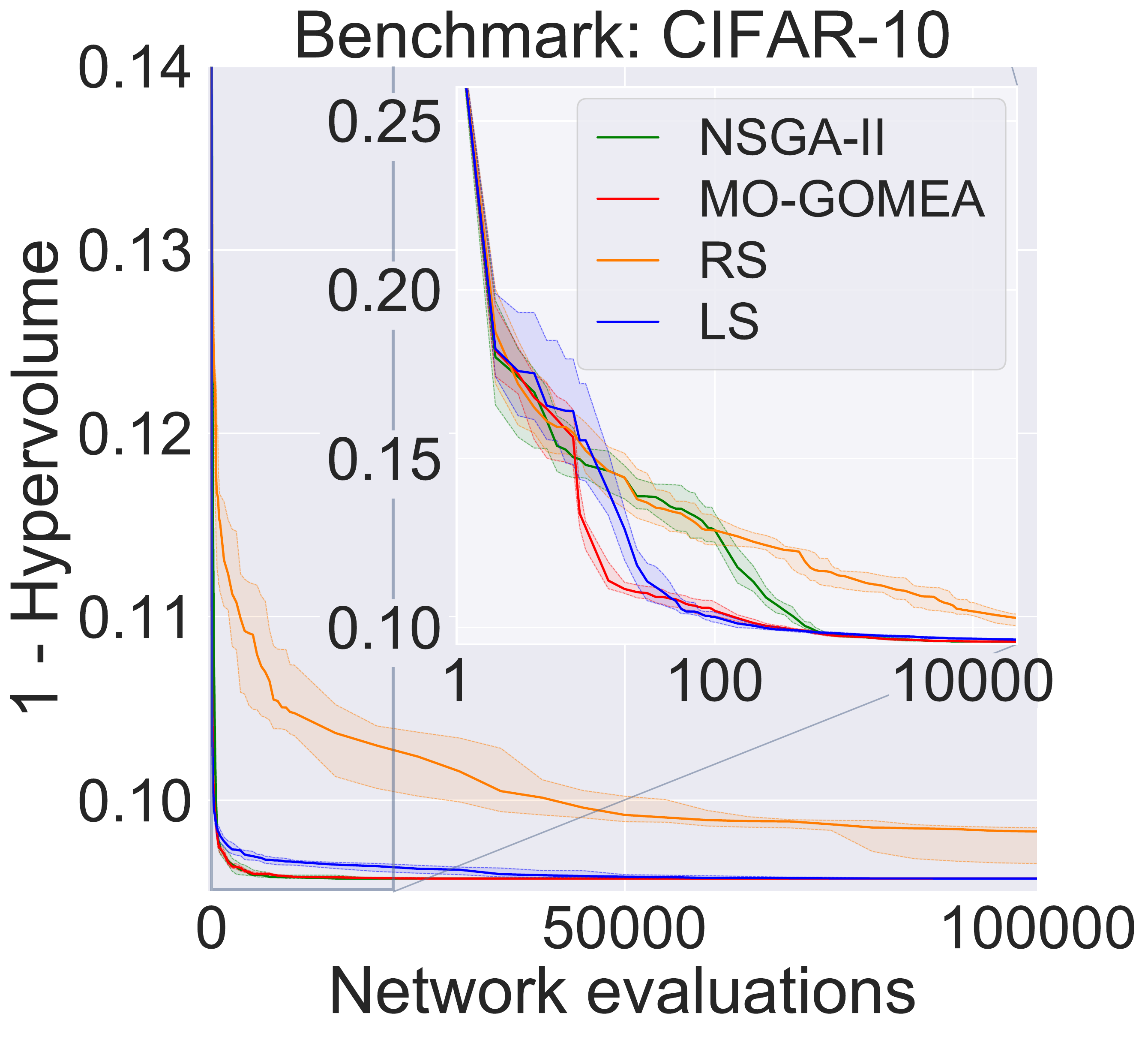}
    \end{subfigure}
    \begin{subfigure}{0.32\linewidth}
        \centering
         \includegraphics[width=\linewidth]{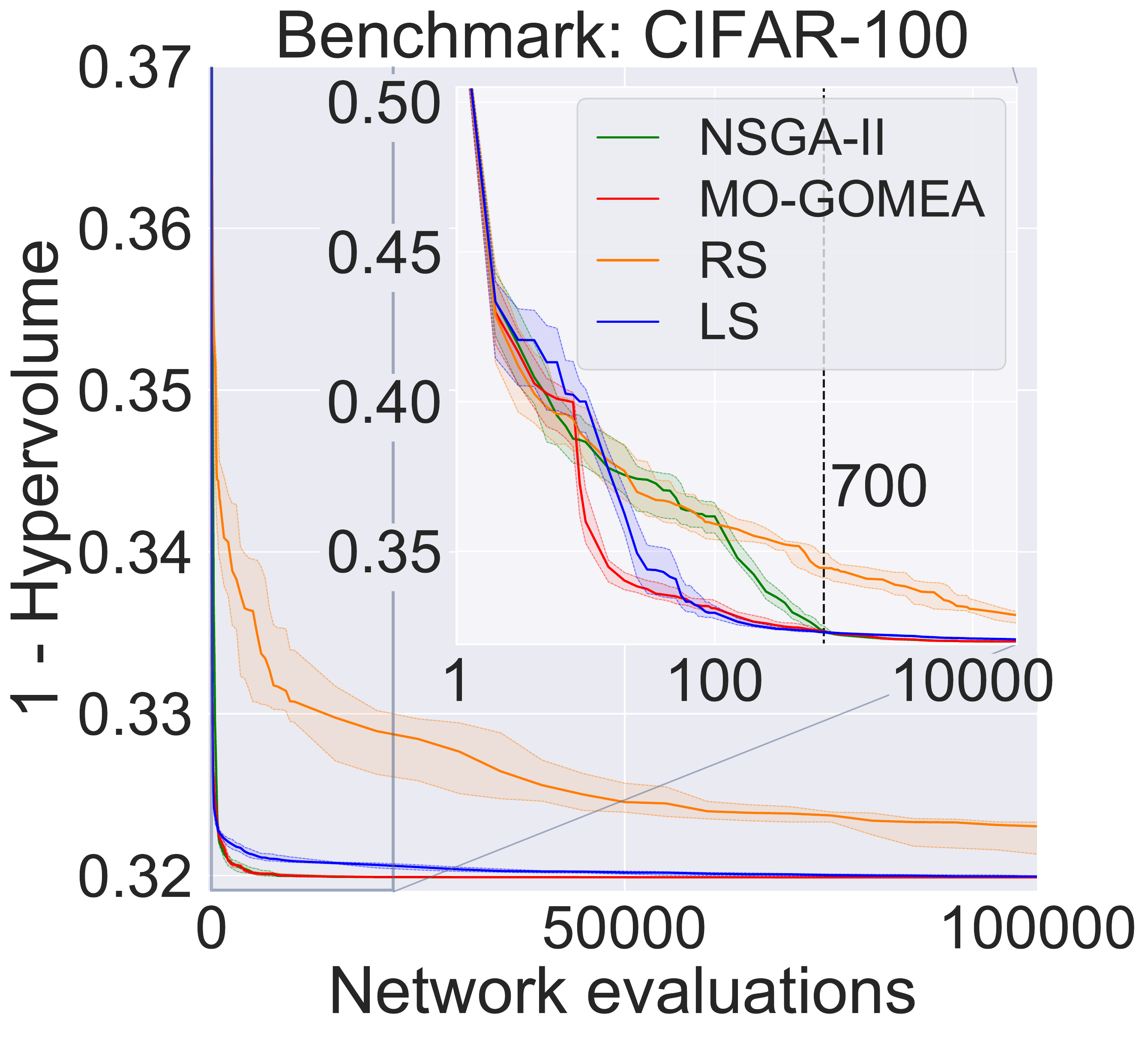}
    \end{subfigure}
    \vspace{-0.3cm}
    \caption{Convergence graphs on NAS-Bench-101 (left), MacroNAS-C10 (middle) and MacroNAS-C100 (right). Medians (solid lines) and 25/75th percentiles (bands) of 30 runs are shown. Note that only the horizontal axes are the same, and are in logarithmic scale in the zoomed-in views (insets).}
    \label{fig:preliminary_exp_hv}
\end{figure}

\begin{figure}[h]
    \vspace{-0.6cm}
    \centering
    \includegraphics[width=\linewidth]{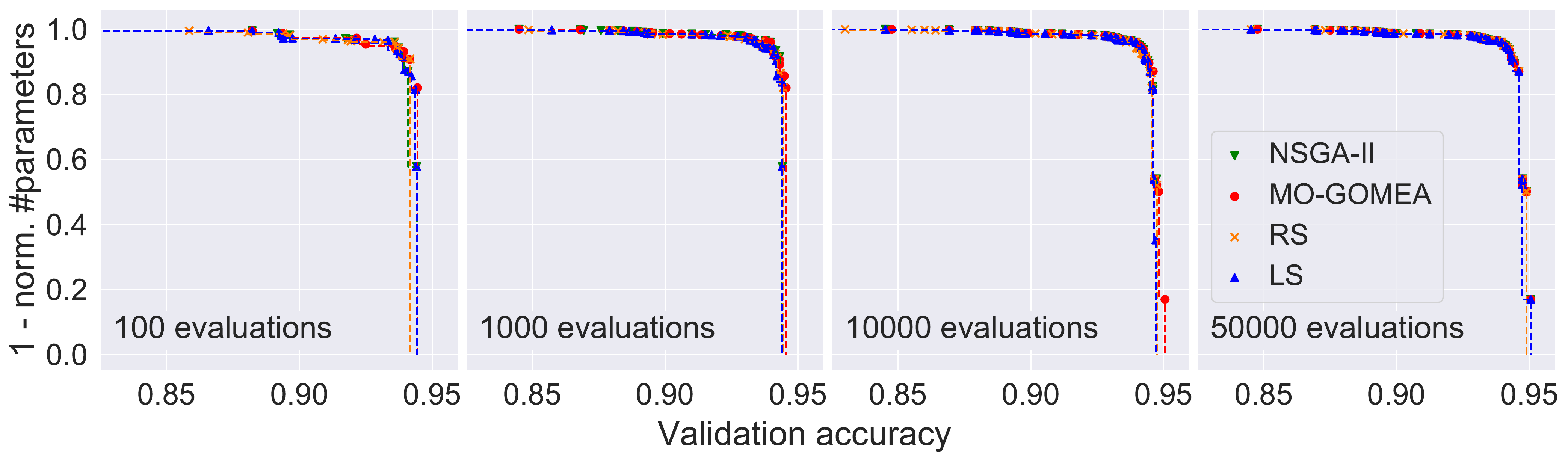}
    \vspace{-0.7cm}
    \caption{Evolving archives for one example run on NAS-Bench-101. 
    }
    \label{fig:NAS101_MO_pareto}
    \vspace{-1.0cm}
\end{figure}

We begin by analyzing the results for NAS-Bench-101 (Fig.~\ref{fig:preliminary_exp_hv} left). For this benchmark, differences among the algorithms are negligible. This can also be seen in Fig.~\ref{fig:NAS101_MO_pareto}, where fronts obtained by the algorithms are similar at different points in time. 
These results are in line with the ones reported for the single-objective experiments conducted in~\cite{Ying2019NAS-Bench-101}: differences among fundamentally different algorithms are relatively small.

Regarding the macro-level search spaces, i.e., on MacroNAS-C10 and Macro\-NAS-C100 (Fig.~\ref{fig:preliminary_exp_hv} middle and right, respectively), a notable difference is found between RS and the other algorithms. There are however also small differences between the EAs and LS. In the zoomed-in views it can be seen that, for the first 100 evaluations, NSGA-II performs as good as RS: this is because the initial population ($n_\textit{pop}=100$) is evaluated. Meanwhile, MO-GOMEA already performed optimization on the small initial populations instantiated by its IMS.

The hypervolume obtained for CIFAR-100 is not as good as the one obtained for CIFAR-10. This is not surprising because CIFAR-100 is considered a harder classification task. On the MacroNAS benchmarks, the best \accval{} for CIFAR-10 is $0.925$, the one for CIFAR-100 is $0.705$. Furthermore, note that the fact that the hypervolumes for NAS-Bench-101 reach better values, is mainly because $f_2$ is different here (based on number of net parameters instead of MMACs). Still, NAS-Bench-101 contains nets with slightly larger \accval{} (yet easily found by all algorithms), up to $0.951$.
Since CIFAR-100 is the harder task, we focus on this task in the next sections.

\vspace{-0.3cm}
\subsection{MacroNAS-C100}\label{sec:exp-benchmarks-mo}

Fig.~\ref{fig:preliminary_exp_hv} (right) shows that, particularly for the EAs, the hypervolume increases rather quickly in the beginning, and not notably afterwards. Covering the entire Pareto front is however hard (e.g., only 21 out of 30 runs of MO-GOMEA cover it within 100,000 evals).
Overall, MO-GOMEA performs best, followed by LS until $\sim 700$ evaluations. At this point, differences between MO-GOMEA, NSGA-II and LS are not significant at the 99\% confidence level.

We remark that what these differences truly mean is discussed in terms of generalization in Sec.~\ref{sec:generalization}. 
RS is clearly the inferior approach.

\begin{figure}[h]
    \vspace{-0.5cm}
    \centering
    \includegraphics[width=0.99\linewidth]{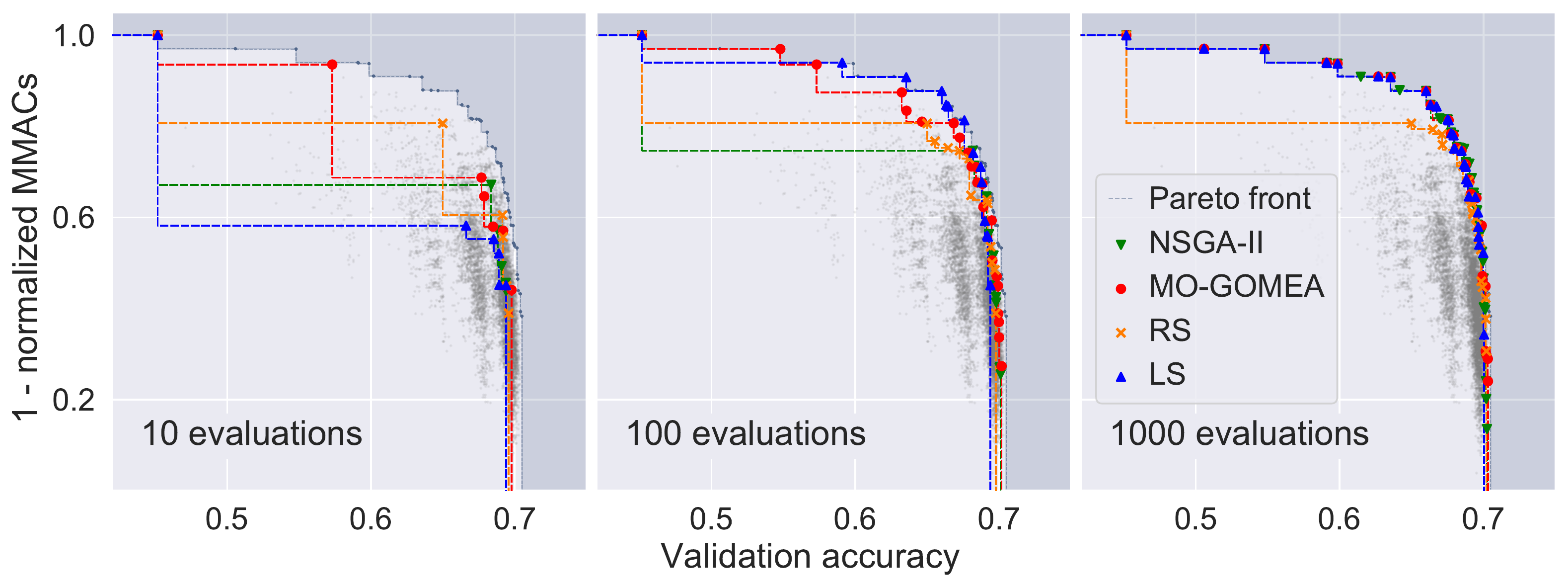}
    \vspace{-0.5cm}
    \caption{Evolving archives for one example run on the MacroNAS-C100 benchmark dataset. Gray points represent all possible architectures.}
    \label{fig:BenchC100_archive_val}
    \vspace{-0.9cm}
\end{figure}

To obtain a better understanding of the search the algorithms perform, we present the fronts discovered for CIFAR-100 at different time steps in Fig.~\ref{fig:BenchC100_archive_val} and Fig.~\ref{fig:BenchC100_largescale_archive_test} (left) (we found very similar results for MacroNAS-C10). The plots for 10 and 100 evaluations confirm the fact that MO-GOMEA and LS are the quickest at approaching the front. Note that the archive at 10 evaluations (Fig.~\ref{fig:BenchC100_archive_val} left) of LS is already obtained during the very first iteration of LS.

\begin{wrapfigure}{r}{.3\linewidth}
    \vspace{-0.7cm}
    \begin{subfigure}{\linewidth}
        \centering
        \includegraphics[width=\linewidth]{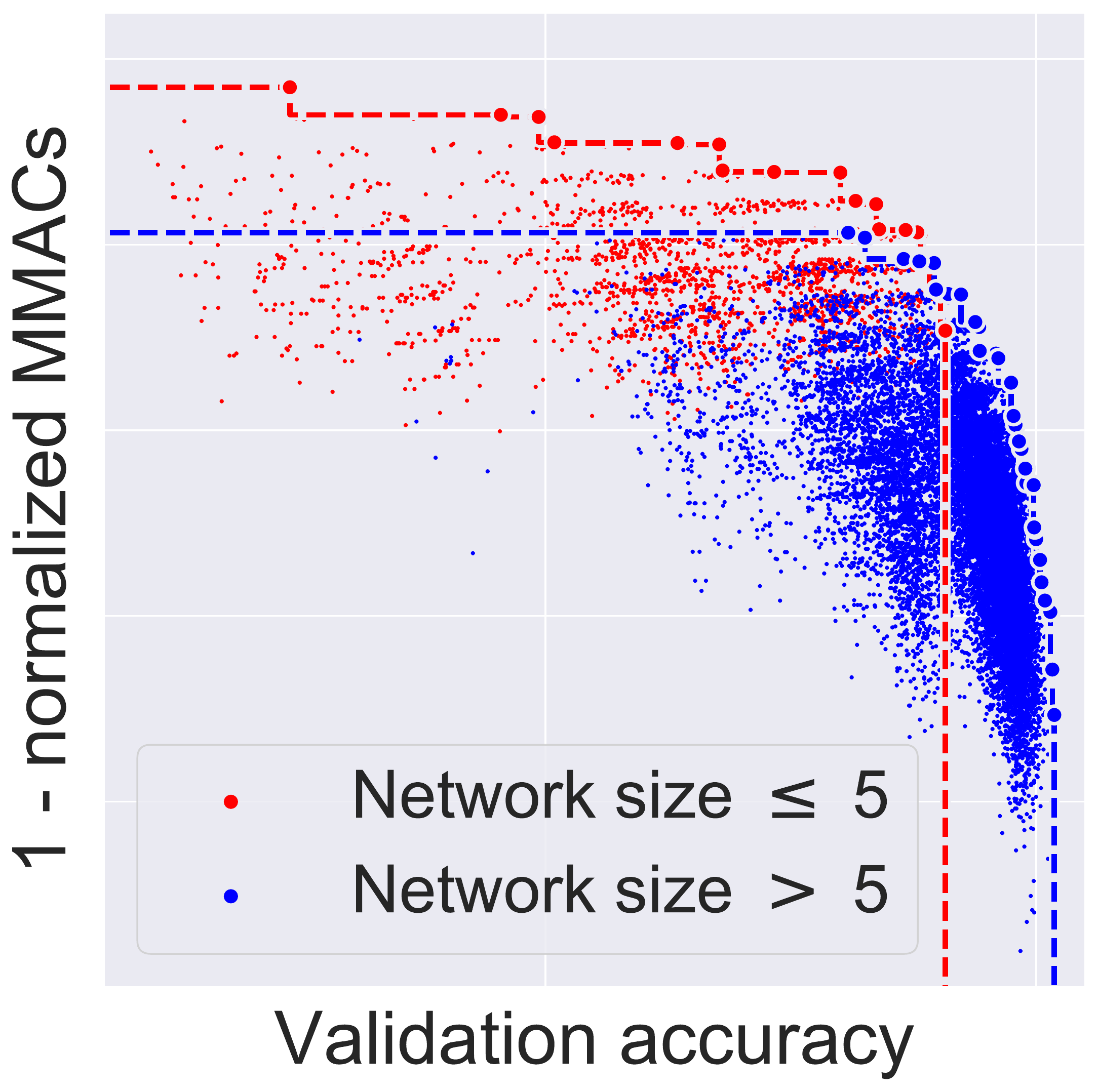}
    \end{subfigure}
    \vspace{-0.3cm}

    \caption{
    Distribution of two classes of nets and their Pareto fronts.
    }
    
    \label{fig:bench_identity_scatter}
    \vspace{-0.8cm}
\end{wrapfigure}

Compared to MO-GOMEA, LS is especially quick at finding the less-densely populated areas in the objective space (low MMACs). 
This is likely because making networks with less MMACs is straightforward, as more layers can be set to identity, which is effectively searched by the LS (with the proper objective weighting) that tries every option in each cell.
Moreover, the EAs need to rely on the encodings present in the initial population, where an identity cell is sampled in position $i$ only with chance $1/|\Omega_i|$. 
As shown in Fig.~\ref{fig:bench_identity_scatter},  evaluating small architectures is necessary to cover the part of the Pareto front containing efficient nets, as the number of nets there is smaller.

It is clear that RS is inferior to the other algorithms, as it struggles to find efficient nets that occur only sparsely in the search space. This issue is due to the fact that RS does not build up from efficient nets and thus solely relies on the way it samples: sampling a net with many identity cell is rare. For example, the probability of sampling the full-identity net is $1/|\Omega_i|^\ell$ (recall that this net is pre-inserted in the archive because uninteresting).

\subsection{Large-scale NAS: CIFAR-100}\label{sec:exp-largescale}

Fig.~\ref{fig:BenchC100_largescale_HV_test} (right; consider only \accval{}, i.e., solid lines) shows how the hypervolume improves over time for the large-scale experiment on  CIFAR-100. MO-GOMEA and LS obtain a good hypervolume similarly quickly, as observed for MacroNAS-C100. NSGA-II converges more slowly than MO-GOMEA and LS, but it is capable of outperforming them in terms of hypervolume towards the end. Nonetheless, both LS and MO-GOMEA do appear to improve again near the end of our total evaluation budget.

\begin{figure}[H]
    \vspace{-0.8cm}
    \centering
    \includegraphics[width=\linewidth]{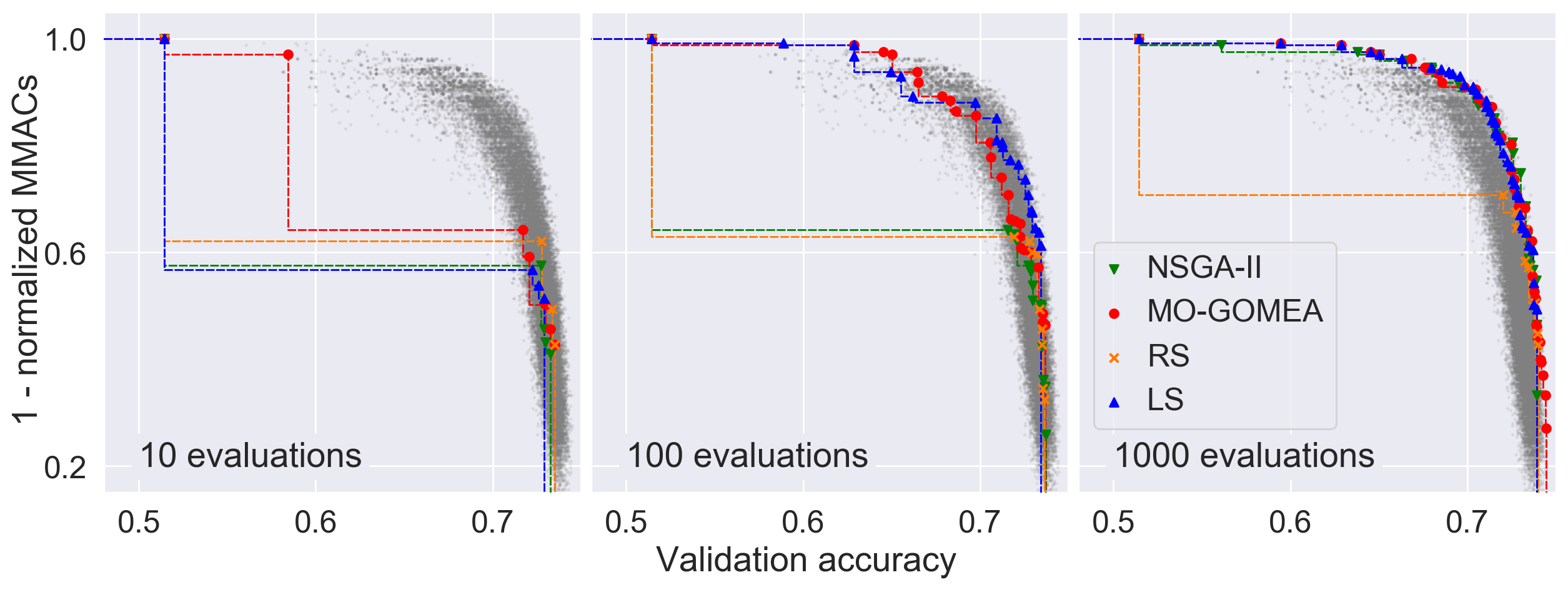}
    \vspace{-0.8cm}
    \caption{Evolving archives for one example run of the large-scale NAS for CIFAR-100. Gray points represent all architectures ever discovered.}
    \label{fig:largescale_archive_val}
    \vspace{-0.5cm}
\end{figure}

\begin{figure}[H]
    \vspace{-0.8cm}
    \centering
    \includegraphics[width=\linewidth]{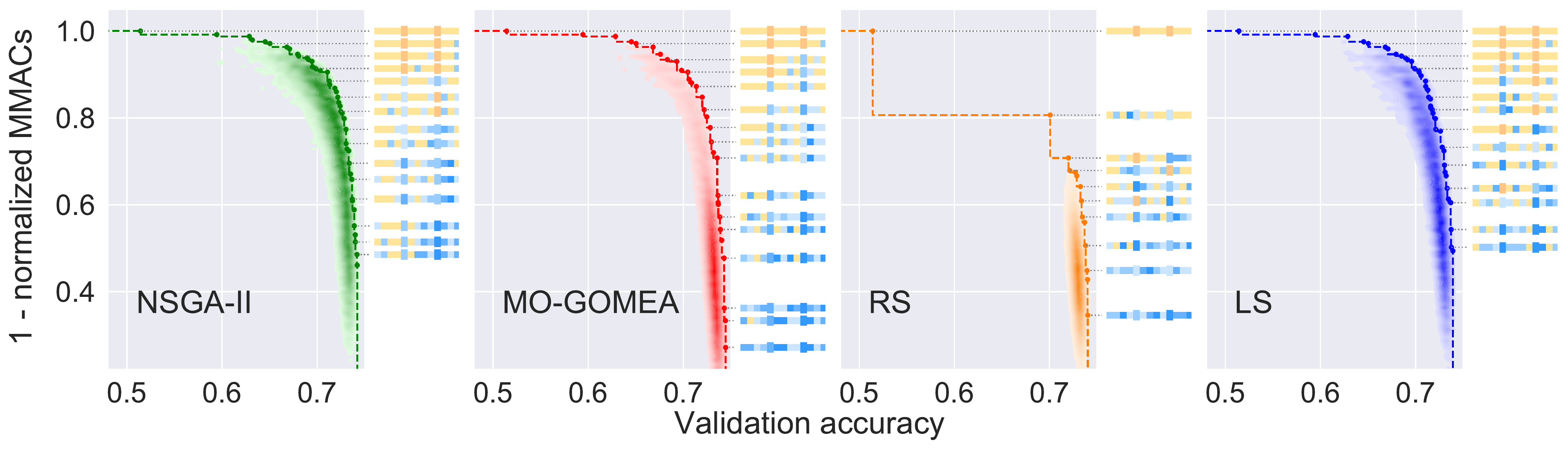}
    \vspace{-0.8cm}
    \caption{Fronts obtained at 2500 evaluations by the search algorithms in one example run on the large-scale experiment on CIFAR-100, with a visual description of the nets. The colored cells correspond to the ones displayed in Figure~\ref{fig:example-search-space}. Heatmaps display the distribution of nets evaluated by the algorithms.}
    \label{fig:largescale_architectures_heatmap}
    \vspace{-0.8cm}
\end{figure}

Fig.~\ref{fig:largescale_archive_val} shows the way the algorithms discover solutions over time, and Fig.~\ref{fig:largescale_architectures_heatmap} shows the density of architectures obtained at 2500 evaluations. Notably, MO-GOMEA discovers nets with the largest \accval{} (bottom-right points, consistent for all 6 runs). This may be because of two reasons. First, MO-GOMEA exploits linkage, which may be useful to recombine complex building blocks and obtain better performing nets. Second, MO-GOMEA restricts recombination to solutions that are similar.
 
\begin{wrapfigure}{r}{.4\linewidth}
\vspace{-0.7cm}
    \centering
    \begin{subfigure}{\linewidth}
      \centering
      \includegraphics[width=\linewidth]{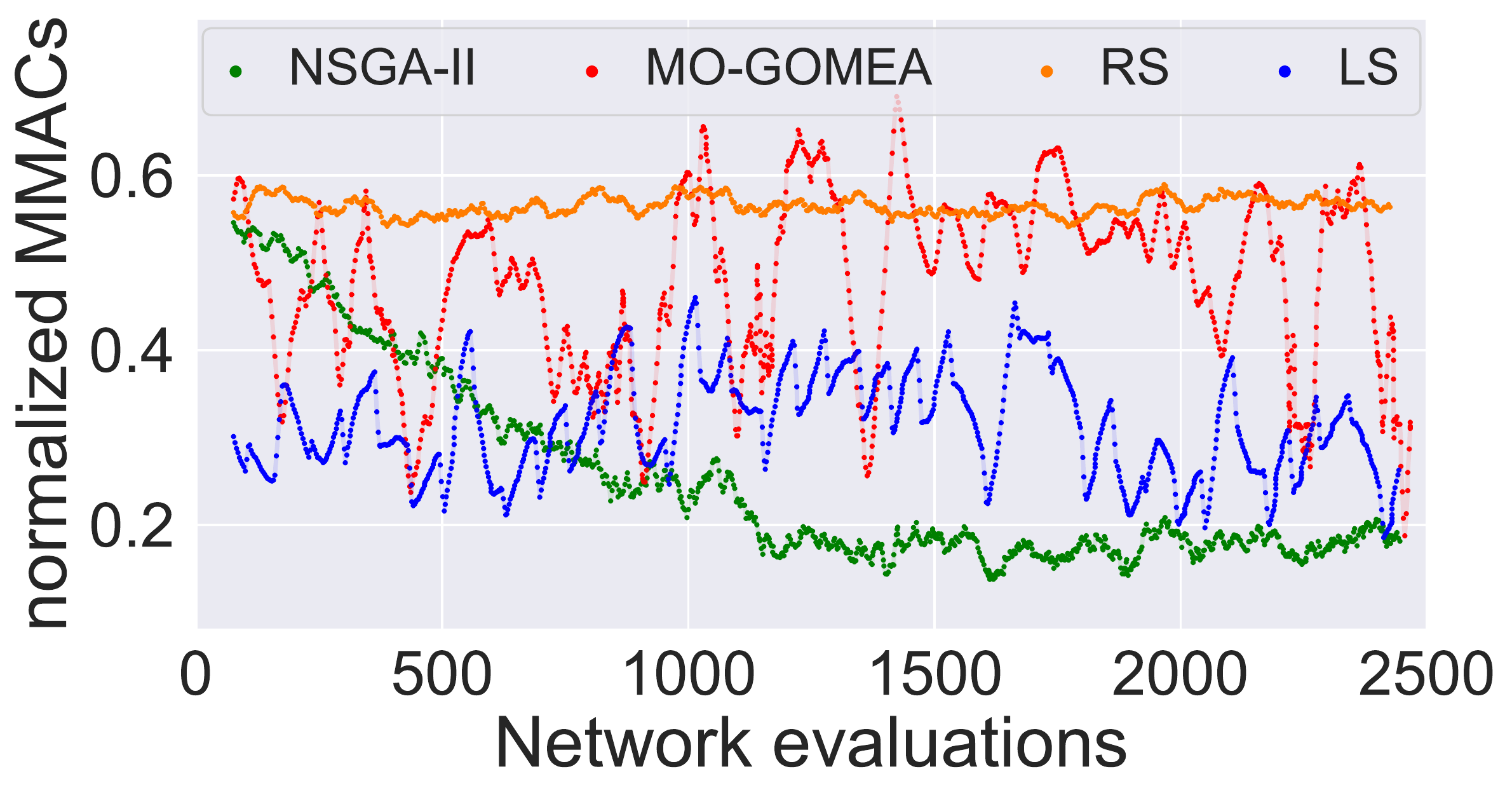}
    \end{subfigure}
    \vspace{-0.3cm}
    \caption{MMACs of evaluated nets throughout one example run, smoothed for readability (moving avg. filter of size $75$).
    }
    \label{fig:MMAC_analysis}
    \vspace{-0.5cm}
\end{wrapfigure}

Regarding RS, from Figs.~\ref{fig:MMAC_analysis}, \ref{fig:largescale_architectures_heatmap} it can be seen that RS mostly samples complex nets. Yet, as RS lacks any mechanism to refine and exploit solutions, it cannot match MO-GOMEA's capability of discovering the most accurate nets. The main issue of RS is, as aforementioned, that it samples efficient nets too rarely.  

NSGA-II exhibits a very interesting behavior: it searches progressively from more complex to simpler architectures (Figs.~\ref{fig:largescale_archive_val}, \ref{fig:MMAC_analysis}). This happens because in the beginning it is more likely to generate complex nets (as RS does), while later on NSGA-II's crowding distance operator steers the search towards the areas where points are most scattered, i.e., where efficient nets are located. Moreover, it is likely that its non-linkage informed operators are less likely to generate the best performing nets.

LS is found to be slightly inferior to the EAs in terms of final hypervolume (Fig.~\ref{fig:BenchC100_largescale_HV_test} right, solid lines), and it does not ultimately obtain the best front (Figs.~\ref{fig:largescale_archive_val}, and~\ref{fig:BenchC100_largescale_archive_test} middle-right).
Having only 6 repetitions, statistical testing results in no significant differences for the hypervolume at the end of the runs (i.e., at 2500 evaluations). 
What we believe is most important is that LS is remarkably quick at obtaining fairly good nets (Fig.~\ref{fig:largescale_archive_val}), with rather uniform spread in terms of trade-offs (Fig.~\ref{fig:MMAC_analysis}). Moreover, LS has no issues similar to RS, in fact, it outperforms it markedly.

\subsection{Generalization}\label{sec:generalization}
Because the validation is used during the search, a risk exists to implicitly start overfitting to it~\cite{bishop2006pattern}. Hence, a second held-out set is used, i.e., given (a front of) best-\accval{} nets, their \acctest{} is measured.

\begin{figure}
    \vspace{-0.5cm}
    \centering
    \begin{subfigure}{0.49\linewidth}
        \centering
        \includegraphics[width=\linewidth]{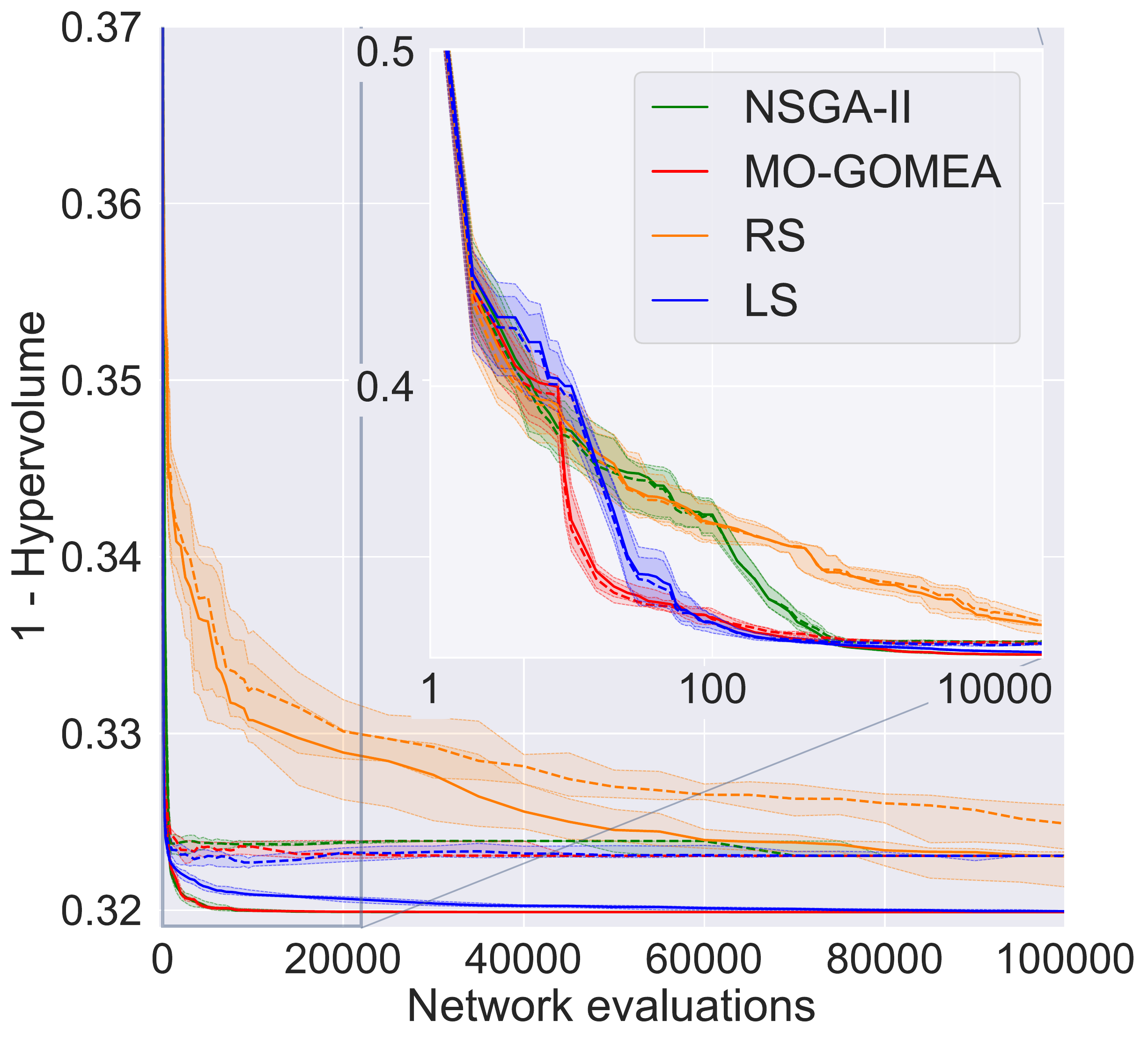}
    \end{subfigure}
        \begin{subfigure}{0.482\linewidth}
        \centering
        \includegraphics[width=\linewidth]{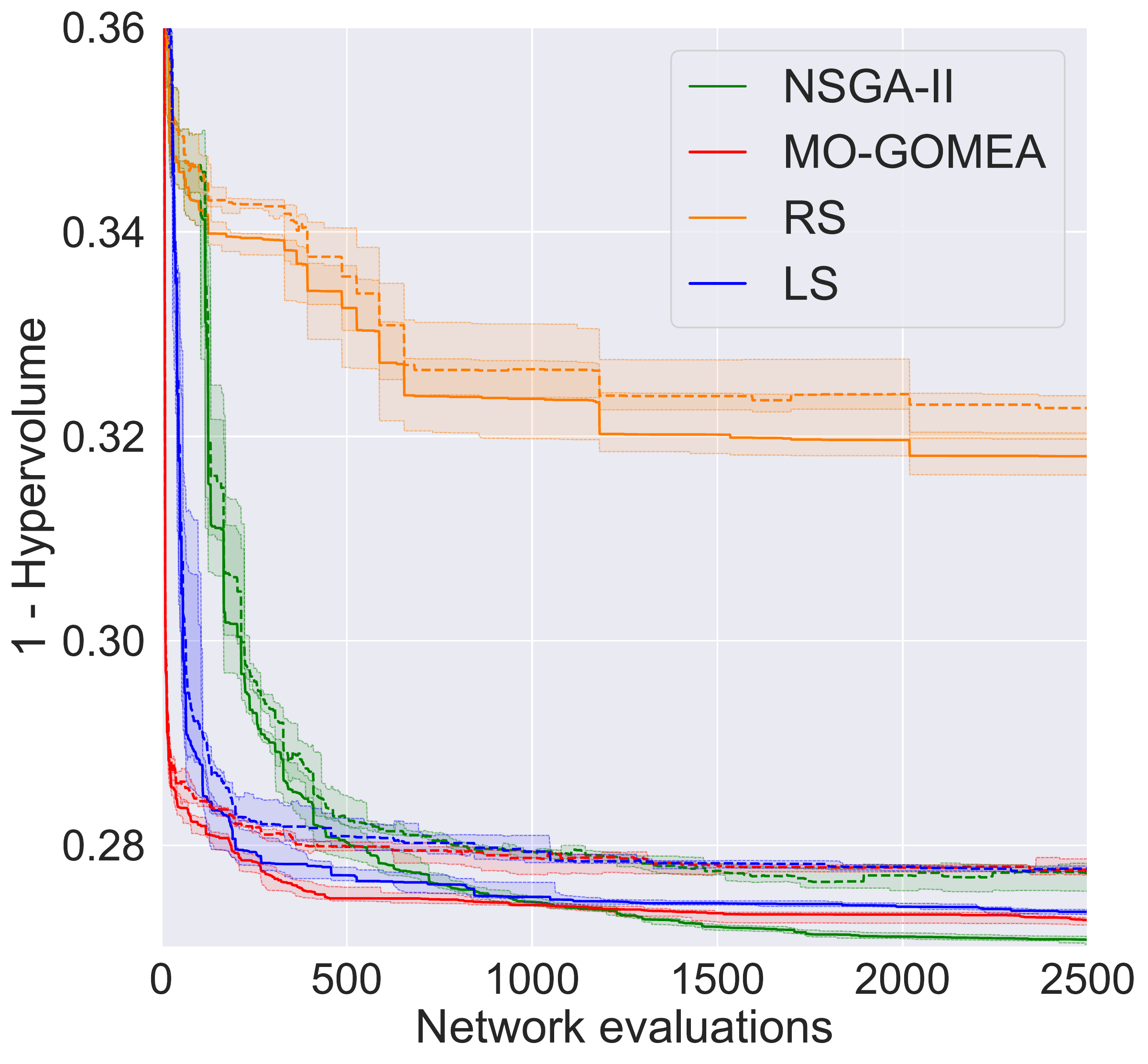}
    \end{subfigure}
    \vspace{-0.3cm}
    \caption{Convergence graphs regarding \accval{} (solid lines) and \acctest{} (dashed lines) on MacroNAS-C100 (left) and on CIFAR-100 on the enlarged search space (right). Medians (lines) and 25/75th percentiles (bands) of respectively 30 and 6 runs are shown for all algorithms. Note the different ranges for the vertical axes and the logarithmic scale for the horizontal axis in the zoomed-in view.}
    \label{fig:BenchC100_largescale_HV_test}
\end{figure}
\begin{figure}[t]
    \vspace{-0.8cm}
    \centering
    \begin{subfigure}{0.523\linewidth}
        \centering
        \includegraphics[width=\linewidth]{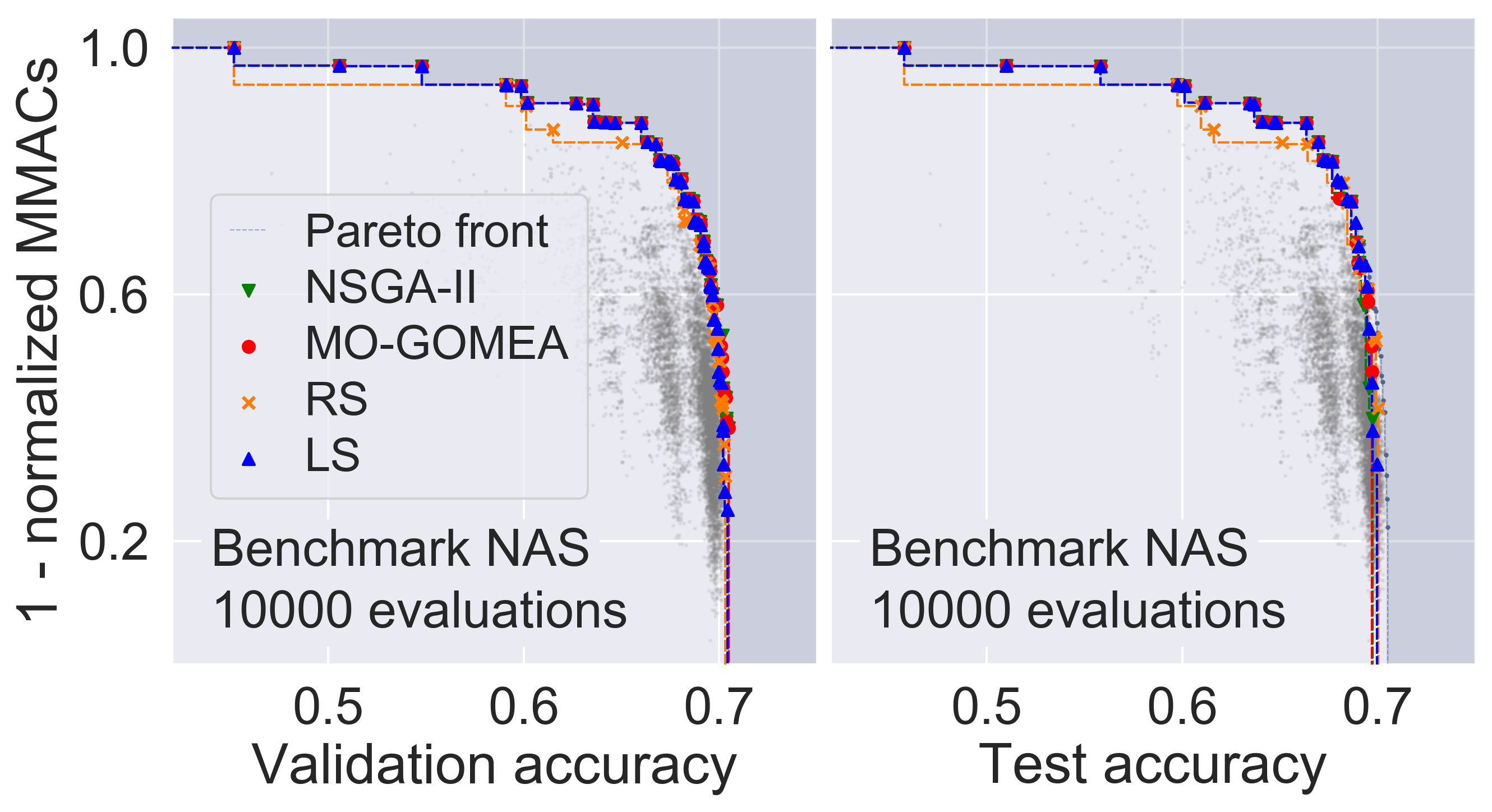}
    \end{subfigure}
    \begin{subfigure}{0.468\linewidth}
        \centering
        \includegraphics[width=\linewidth]{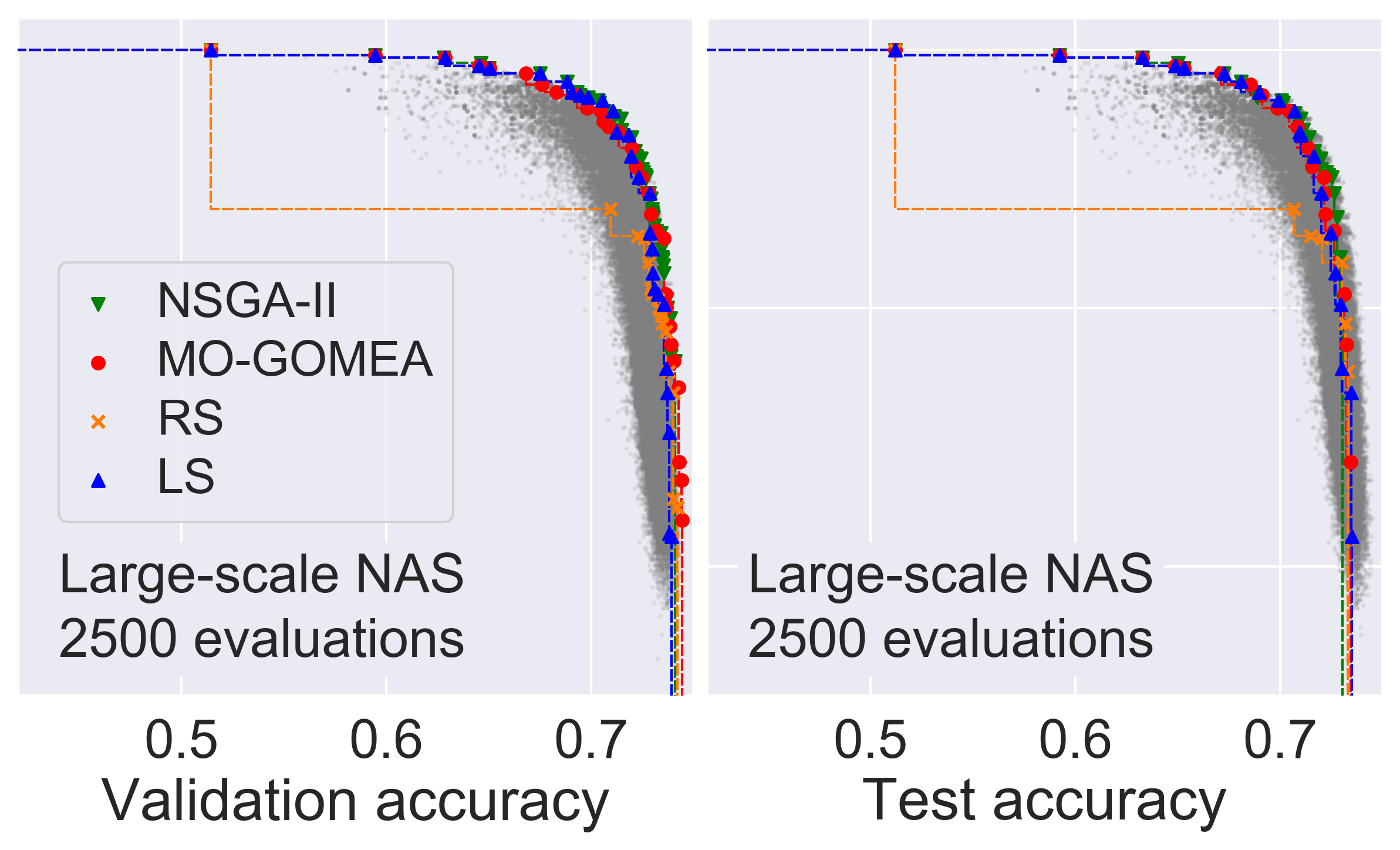}
    \end{subfigure}
    \vspace{-0.3cm}
    \caption{Validation and test archives for 10,000 evaluations on MacroNAS-C100 (resp. first and second from the left) and 2500 evaluations on large-scale NAS (resp. second and first from the right). Archives are from one example run.}
    \label{fig:BenchC100_largescale_archive_test}
    \vspace{-0.3cm}
\end{figure}

As can be seen from Figs.~\ref{fig:BenchC100_largescale_HV_test} (left) and \ref{fig:BenchC100_largescale_archive_test} (left-most two), on MacroNAS-C100 RS clearly remains the worst performing algorithm when considering generalization to the test set. However, MO-GOMEA and NSGA-II lose the lead to LS, between 100 and 18,000 evaluations. This is a consequence of the aforementioned phenomenon: despite the fact that nets are not trained on the set of data on which \accval{} is measured, \accval{} is used to select what nets are best, hence the EAs' better search translates to more overfitting to this set. LS is slightly less powerful at searching (worse hypervolume in terms of \accval{}), yet the nets generalize well to the test set. Crucially, there are no significant differences (at 99\% confidence) among LS and the EAs at 100,000 evaluations.



The findings for the enlarged search space reflect the ones found for Macro\-NAS-C100: the difference in performance between LS and the EAs becomes smaller when considering \acctest{} (see Fig.~\ref{fig:BenchC100_largescale_HV_test} right). Fig.~\ref{fig:BenchC100_largescale_archive_test} (right-most two) displays this especially well, as the most accurate nets discovered by MO-GOMEA turn out to be less accurate when tested again. Again, no statistical differences are observed at the end of the runs between LS and the EAs. 


\vspace{-1.1cm}
\section{Discussion} \label{sec:discussion} 

While several works compare algorithms in terms of time, we preferred a comparison in terms of number of evaluations. Using a time budget is reasonable for practical applications, but has its limitations: particularly, time-based experiments are influenced by hyper-parameter choices (e.g., network training budgets), and can put some algorithms at disadvantage (e.g., see~\cite{Real2017LargeScaleEvo} for RS). 

In our macro-level comparisons, we systematically found that RS performed much worse than the other algorithms. This boiled down to the fact that RS has a very low probability of sampling efficient nets. This simple result is important when considering RS as a baseline to benchmark against: if the sampling process of RS is not well-aligned with the desired objectives, RS can hardly beat an algorithm that includes optimization, even if as simple as LS. Note that on NAS-Bench-101, where the encoding and sampling process is such that small nets are not extremely rare, RS performed similar to the other algorithms. Carefully setting RS to sample solutions in a way that is well-aligned with the objectives may be non-trivial, and in fact harder than using LS.

We showed that the fact that different EAs include different search mechanisms, such as linkage learning and cluster-restricted mating for MO-GOMEA, and non-dominated sorting and crowding distance for NSGA-II, does lead to notable differences in what nets are found at what point in time. This is important to consider in practical applications where the evaluation budget is limited. LS was found to be competitive in this respect, as it quickly obtained a rather uniform scattering of nets with different trade-offs. Moreover, LS proved capable of ultimately competing very well when generalization is accounted for: the capability of SotA EAs to optimize at a very refined level with respect to \accval{} does not always translate to better \acctest{} due to implicit overfitting.

\section{Conclusion}
With several experiments, we have shown that a simple, parameter-less local search algorithm can be a competitive baseline for NAS. We found that the proposed local search algorithm competes with state-of-the-art EAs, even up to thousands of evaluated architectures, in a multi-objective setting. 
By a greedy single-variable search mechanism and a random scalarization of the objectives, LS discovers nets with rather uniform trade-offs between accuracy and complexity in few evaluations. Performance gaps with EAs evident at search time are often lost when generalization is assessed.
Importantly, we found that local search outperforms random search consistently and markedly. We therefore recommend to adopt local search as a baseline for NAS.





\section*{Acknowledgments}
This work is part of the research programme Commit2Data with (project \#628.011.012), which is financed by the Dutch Research Council (NWO), and part of a project (\#15198) that is included in the research program Technology for Oncology, which is financed by the Netherlands Organization for Scientific Research (NWO), the Dutch Cancer Society (KWF), the TKI Life Sciences \& Health, Asolutions, Brocacef, and Cancer Health Coach. 
\bibliographystyle{splncs04}
\bibliography{references}

\begin{thebibliography}{10}
\providecommand{\url}[1]{\texttt{#1}}
\providecommand{\urlprefix}{URL }
\providecommand{\doi}[1]{https://doi.org/#1}

\bibitem{Bender2018UnderstandingAS}
Bender, G., Kindermans, P.J., Zoph, B., Vasudevan, V., Le, Q.: Understanding
  and simplifying one-shot architecture search. In: Proceedings of the 35th
  International Conference on Machine Learning. Proceedings of Machine Learning
  Research, vol.~80, pp. 550--559. PMLR, Stockholmsmässan, Stockholm Sweden
  (10--15 Jul 2018)

\bibitem{bishop2006pattern}
Bishop, C.M.: Pattern Recognition and Machine Learning. springer (2006)

\bibitem{bosman2016expanding}
Bosman, P.A.N., Luong, N.H., Thierens, D.: Expanding from discrete {C}artesian
  to permutation gene-pool optimal mixing evolutionary algorithms. In:
  Proceedings of the Genetic and Evolutionary Computation Conference. pp.
  637--644 (2016)

\bibitem{Bosman2010DistanceMetric}
Bosman, P.A.: The anticipated mean shift and cluster registration in
  mixture-based edas for multi-objective optimization. In: Proceedings of the
  Genetic and Evolutionary Computation Conference. p. 351–358. GECCO ’10,
  Association for Computing Machinery, New York, NY, USA (2010).
  \doi{10.1145/1830483.1830549}, \url{https://doi.org/10.1145/1830483.1830549}

\bibitem{cai2018proxylessnas}
Cai, H., Zhu, L., Han, S.: Proxylessnas: Direct neural architecture search on
  target task and hardware. arXiv preprint arXiv:1812.00332  (2018)

\bibitem{Chu2019MOREMNAS}
Chu, X., Zhang, B., Xu, R., Ma, H.: Multi-objective reinforced evolution in
  mobile neural architecture search. arXiv preprint arXiv:1901.01074  (2019)

\bibitem{Deb2002NSGA-II}
{Deb}, K., {Pratap}, A., {Agarwal}, S., {Meyarivan}, T.: A fast and elitist
  multiobjective genetic algorithm: {NSGA-II}. IEEE Transactions on
  Evolutionary Computation  \textbf{6}(2),  182--197 (April 2002)

\bibitem{dong2020nasbench201}
Dong, X., Yang, Y.: {NAS-Bench-201}: Extending the scope of reproducible neural
  architecture search. arXiv preprint arXiv:2001.00326  (2020)

\bibitem{Elsken2018LEMONADE}
Elsken, T., Metzen, J.H., Hutter, F.: Efficient multi-objective neural
  architecture search via lamarckian evolution. arXiv preprint arXiv:1804.09081
   (2018)

\bibitem{Elsken2018NASSurvey}
Elsken, T., Metzen, J.H., Hutter, F.: Neural architecture search: A survey.
  Journal of Machine Learning Research  \textbf{20}(55),  1--21 (2019)

\bibitem{guo2019single}
Guo, Z., Zhang, X., Mu, H., Heng, W., Liu, Z., Wei, Y., Sun, J.: Single path
  one-shot neural architecture search with uniform sampling. arXiv preprint
  arXiv:1904.00420  (2019)

\bibitem{he2016deep}
He, K., Zhang, X., Ren, S., Sun, J.: Deep residual learning for image
  recognition. In: Proceedings of the IEEE conference on computer vision and
  pattern recognition. pp. 770--778 (2016)

\bibitem{izmailov2018averaging}
Izmailov, P., Podoprikhin, D., Garipov, T., Vetrov, D., Wilson, A.G.: Averaging
  weights leads to wider optima and better generalization. arXiv preprint
  arXiv:1803.05407  (2018)

\bibitem{jouppi2017datacenter}
Jouppi, N.P., Young, C., Patil, N., Patterson, D., Agrawal, G., Bajwa, R.,
  Bates, S., Bhatia, S., Boden, N., Borchers, A., et~al.: In-datacenter
  performance analysis of a tensor processing unit. In: Proceedings of the 44th
  Annual International Symposium on Computer Architecture. pp. 1--12 (2017)

\bibitem{Kim2017NEMO}
Kim, Y.H., Reddy, B., Yun, S., Seo, C.: Nemo: Neuro-evolution with
  multiobjective optimization of deep neural network for speed and accuracy.
  In: JMLR: Workshop and Conference Proceedings. vol.~1, pp.~1--8 (2017)

\bibitem{kitchin2014data}
Kitchin, R.: The data revolution: {B}ig data, open data, data infrastructures
  and their consequences. Sage (2014)

\bibitem{Lu2018NSGA-Net}
Lu, Z., Whalen, I., Boddeti, V., Dhebar, Y., Deb, K., Goodman, E., Banzhaf, W.:
  {NSGA}-{Net}: Neural architecture search using multi-objective genetic
  algorithm. In: Proceedings of the Genetic and Evolutionary Computation
  Conference. p. 419–427. GECCO 2019, Association for Computing Machinery,
  New York, NY, USA (2019)

\bibitem{Luong2018MO-GOMEA}
Luong, N.H., Poutré, H.L., Bosman, P.A.N.: Multi-objective gene-pool optimal
  mixing evolutionary algorithm with the interleaved multi-start scheme. Swarm
  and Evolutionary Computation  \textbf{40},  238 -- 254 (2018)

\bibitem{miikkulainen2019codeepneat}
Miikkulainen, R., Liang, J., Meyerson, E., Rawal, A., Fink, D., Francon, O.,
  Raju, B., Shahrzad, H., Navruzyan, A., Duffy, N., et~al.: Evolving deep
  neural networks. In: Artificial Intelligence in the Age of Neural Networks
  and Brain Computing, pp. 293--312. Elsevier (2019)

\bibitem{Pelikan2006Scalable}
Pelikan, M., Sastry, K., Cant\'{u}-Paz, E.: Scalable Optimization via
  Probabilistic Modeling: From Algorithms to Applications (Studies in
  Computational Intelligence). Springer-Verlag, Berlin, Heidelberg (2006)

\bibitem{radford2019language}
Radford, A., Wu, J., Child, R., Luan, D., Amodei, D., Sutskever, I.: Language
  models are unsupervised multitask learners. OpenAI Blog  \textbf{1}(8), ~9
  (2019)

\bibitem{Real2017LargeScaleEvo}
Real, E., Moore, S., Selle, A., Saxena, S., Suematsu, Y.L., Tan, J., Le, Q.V.,
  Kurakin, A.: Large-scale evolution of image classifiers. In: Proceedings of
  the 34th International Conference on Machine Learning - Volume 70. p.
  2902–2911. ICML 2017, JMLR.org (2017)

\bibitem{russakovsky2015imagenet}
Russakovsky, O., Deng, J., Su, H., Krause, J., Satheesh, S., Ma, S., Huang, Z.,
  Karpathy, A., Khosla, A., Bernstein, M., et~al.: Imagenet large scale visual
  recognition challenge. International Journal of Computer Vision
  \textbf{115}(3),  211--252 (2015)

\bibitem{sandler2018mobilenetv2}
Sandler, M., Howard, A., Zhu, M., Zhmoginov, A., Chen, L.C.: Mobilenetv2:
  Inverted residuals and linear bottlenecks. In: Proceedings of the IEEE
  conference on computer vision and pattern recognition. pp. 4510--4520 (2018)

\bibitem{Yu2020Evaluating}
Sciuto, C., Yu, K., Jaggi, M., Musat, C., Salzmann, M.: Evaluating the search
  phase of neural architecture search. arXiv preprint arXiv:1902.08142  (2019)

\bibitem{senior2020deepmind_protein}
Senior, A.W., Evans, R., Jumper, J., Kirkpatrick, J., Sifre, L., Green, T.,
  Qin, C., {\v{Z}}{\'\i}dek, A., Nelson, A.W., Bridgland, A., et~al.: Improved
  protein structure prediction using potentials from deep learning. Nature
  pp.~1--5 (2020)

\bibitem{silver2017masteringGo}
Silver, D., Schrittwieser, J., Simonyan, K., Antonoglou, I., Huang, A., Guez,
  A., Hubert, T., Baker, L., Lai, M., Bolton, A., et~al.: Mastering the game of
  {Go} without human knowledge. Nature  \textbf{550}(7676),  354--359 (2017)

\bibitem{sze2017efficient}
Sze, V., Chen, Y.H., Yang, T.J., Emer, J.S.: Efficient processing of deep
  neural networks: {A} tutorial and survey. Proceedings of the IEEE
  \textbf{105}(12),  2295--2329 (2017)

\bibitem{tan2019mnasnet}
Tan, M., Chen, B., Pang, R., Vasudevan, V., Sandler, M., Howard, A., Le, Q.V.:
  {MnasNet}: Platform-aware neural architecture search for mobile. In:
  Proceedings of the IEEE Conference on Computer Vision and Pattern
  Recognition. pp. 2820--2828 (2019)

\bibitem{Thierens2011OptMixing}
Thierens, D., Bosman, P.A.N.: Optimal mixing evolutionary algorithms. In:
  Proceedings of the Genetic and Evolutionary Computation Conference. p.
  617–624. GECCO 2011, Association for Computing Machinery (2011)

\bibitem{alphastarblog}
Vinyals, O., Babuschkin, I., Chung, J., Mathieu, M., Jaderberg, M., Czarnecki,
  W., Dudzik, A., Huang, A., Georgiev, P., Powell, R., Ewalds, T., Horgan, D.,
  Kroiss, M., Danihelka, I., Agapiou, J., Oh, J., Dalibard, V., Choi, D.,
  Sifre, L., Sulsky, Y., Vezhnevets, S., Molloy, J., Cai, T., Budden, D.,
  Paine, T., Gulcehre, C., Wang, Z., Pfaff, T., Pohlen, T., Yogatama, D.,
  Cohen, J., McKinney, K., Smith, O., Schaul, T., Lillicrap, T., Apps, C.,
  Kavukcuoglu, K., Hassabis, D., Silver, D.: {AlphaStar: Mastering the
  Real-Time Strategy Game StarCraft II}.
  \url{https://deepmind.com/blog/alphastar-mastering-real-time-strategy-game-starcraft-ii/}
  (2019)

\bibitem{white2020local}
White, C., Nolen, S., Savani, Y.: Local search is state of the art for {NAS}
  benchmarks. arXiv preprint arXiv:2005.02960  (2020)

\bibitem{wistuba2019nassurvey}
Wistuba, M., Rawat, A., Pedapati, T.: A survey on neural architecture search.
  arXiv preprint arXiv:1905.01392  (2019)

\bibitem{wong2018transfer}
Wong, C., Houlsby, N., Lu, Y., Gesmundo, A.: Transfer learning with {Neural}
  {AutoML}. In: Bengio, S., Wallach, H., Larochelle, H., Grauman, K.,
  Cesa-Bianchi, N., Garnett, R. (eds.) Advances in Neural Information
  Processing Systems 31, pp. 8356--8365. Curran Associates, Inc. (2018)

\bibitem{Yang2020NASishard}
Yang, A., Esperan{\c{c}}a, P.M., Carlucci, F.M.: {NAS} evaluation is
  frustratingly hard. arXiv preprint arXiv:1912.12522  (2019)

\bibitem{Ying2019NAS-Bench-101}
Ying, C., Klein, A., Real, E., Christiansen, E., Murphy, K., Hutter, F.:
  {NAS-Bench-101}: Towards reproducible neural architecture search. arXiv
  preprint arXiv:1902.09635  (2019)

\bibitem{zoph2018learning}
Zoph, B., Vasudevan, V., Shlens, J., Le, Q.V.: Learning transferable
  architectures for scalable image recognition. In: Proceedings of the IEEE
  conference on computer vision and pattern recognition. pp. 8697--8710 (2018)

\end{thebibliography}


\clearpage
\appendix

\section*{Appendix}


In this section we describe in detail the procedure we used to create the benchmark datasets (for CIFAR-10 and CIFAR-100), and the changes introduced for the large-scale NAS experiment (of Sec.~\ref{sec:exp-largescale}).


\subsection*{Design of the MacroNAS benchmark dataset}

We consider feed-forward CNNs, consisting of 17 sequentially connected cells. Similarly to \cite{Ying2019NAS-Bench-101}, each net includes two auxiliary components: the stem convolution prior to the first cell, and a classifier after the last cell, which maps the output of the last cell to a prediction score for each class. The stem convolution is implemented by a $3\times3$ convolutional layer which converts the input image to 32 feature maps. The classifier is implemented by a $1\times1$ convolutional layer, followed by global average pooling and the final linear layer. 
Fig.~\ref{fig:example-search-space} shows an example of our NAS setting. 

Similarly to~\cite{Ying2019NAS-Bench-101}, we pre-set certain positions to ensure a feasible memory consumption for all nets. Specifically, cells at positions 5, 10 and 15 (\emph{reduction} cells) reduce the spatial input dimensionality and increase the number of channels: for an input of size $D\times H\times W$ ($D$ is number of feature maps, $H$ and $W$ are the spatial dimensions) they return an output of size $2D\times H/2 \times W/2$.
This is implemented by means of a $1\times1$ convolutional layer, followed by max-pooling. The remaining $14$ positions are subjected to search with cells that preserve spatial dimensionality (i.e., \emph{normal} cells).

\begin{figure}[H]
    \vspace{-0.5cm}
    \centering
    \begin{subfigure}{0.65\linewidth}
        \centering
        \includegraphics[width=\linewidth]{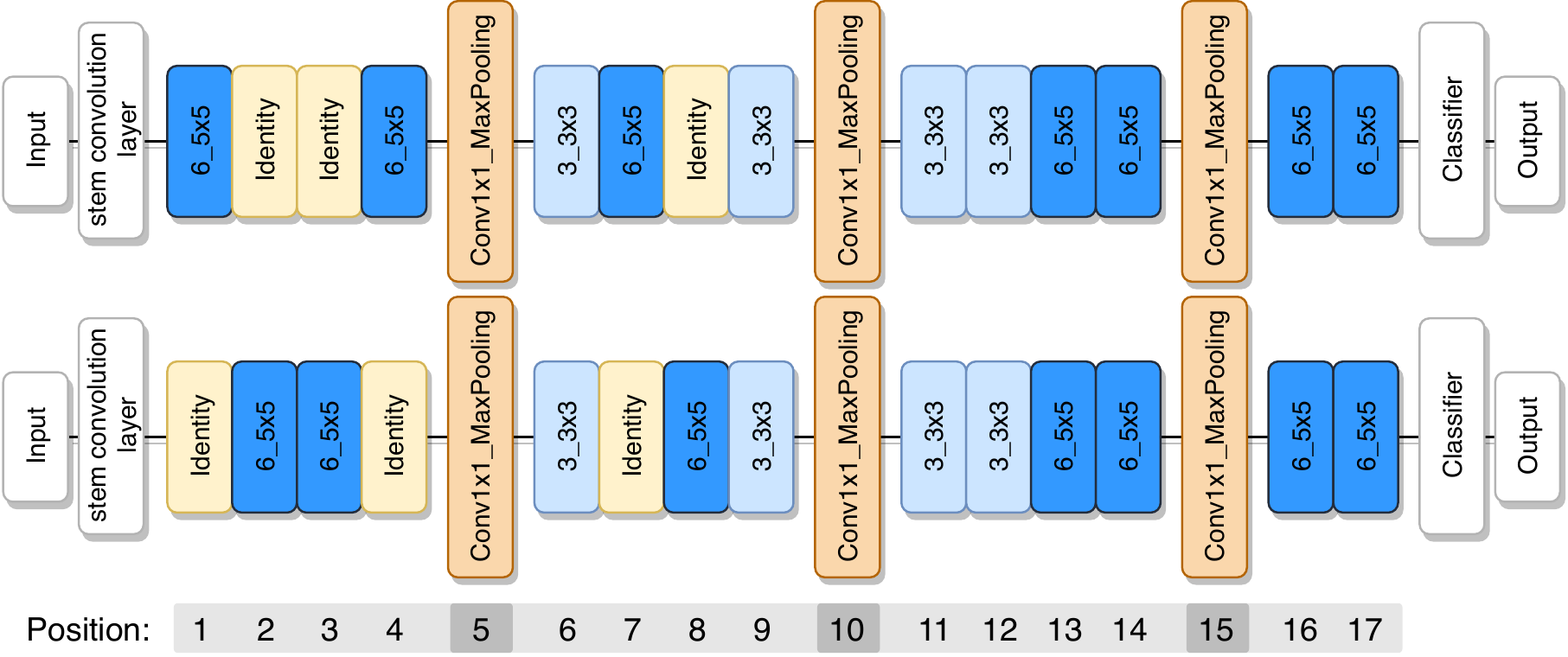}
    \end{subfigure}
    \hspace{0.2cm}
    \begin{subfigure}{0.3\linewidth}
        \centering
        \includegraphics[width=\linewidth]{img/example_network_legend_V2.pdf}
    \end{subfigure}
    \vspace{-0.2cm}
    \caption{Two computationally equivalent networks (left) and legend of cell types (right)}
    \label{fig:example_networks_equivalent}
    \vspace{-0.5cm}
\end{figure}

As options for the searchable cells, we use differently parameterized inverted bottleneck convolutional blocks (MBConv) \cite{sandler2018mobilenetv2}, which are widely used in NAS and known for their efficiency \cite{guo2019single, cai2018proxylessnas, tan2019mnasnet}. We allow a total of 3 cell types: MBConv with expansion factor 3 and kernel size $3$; MBConv with expansion factor 6 and kernel size $5$; and the \emph{identity} cell. The latter option performs no operation and effectively acts as a placeholder to represent smaller nets, i.e., nets with fewer cells. If two nets are identical except for the position of \emph{identity} cells in-between reduction cells (i.e. from positions $1$ to $4$, $6$ to $9$, and $11$ to $14$), they are computationally equivalent (an example is shown in Fig. \ref{fig:example_networks_equivalent}).


\begin{figure}
    \vspace{-0.3cm}
    \centering
    \includegraphics[width=0.9\linewidth]{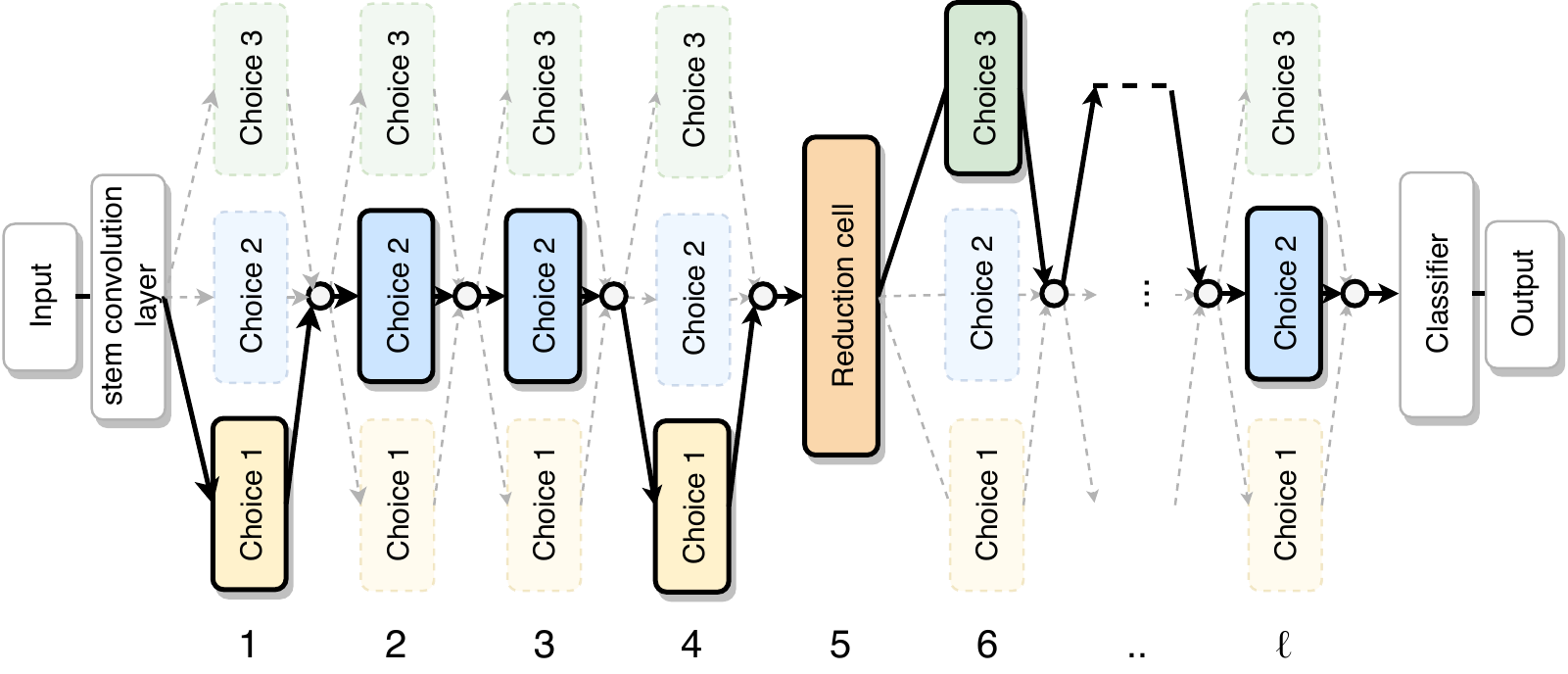}
    \vspace{-0.2cm}
    \caption{Graphical description of the \emph{supernet} with a randomly sampled path highlighted. In our case, the choices correspond to the cells of Fig. \ref{fig:example_networks_equivalent}.}
    \label{fig:supernet}
    \vspace{-1cm}
\end{figure}

To evaluate all nets in a feasible time, we relied on an approach that approximates training from scratch: the one-shot NAS~\cite{guo2019single, Bender2018UnderstandingAS}. The approach trains a \emph{supernet}, i.e., a massive net that contains all possible nets of the search space as sub-nets. We follow the training procedure described in \cite{guo2019single}: for each batch of training samples, one path, i.e., a sequence of cells that represents one of the possible architectures (see Fig.~\ref{fig:supernet}), is sampled from the supernet, and weights in the cells that belong to that path are trained. We use standard training settings: cross-entropy loss and stochastic gradient descent. We use stochastic weight averaging (SWA)~\cite{izmailov2018averaging} for robustness.
For both CIFAR-10/100, we use a stratified data split as done in~\cite{Ying2019NAS-Bench-101}: 40K samples for training, 10K for validation, 10K for testing. Further details on the training process are summarized in Table \ref{tab:training_hyperparameters}.

To evaluate a single architecture, we follow the approach also described in \cite{guo2019single}: 1) Take the weights from the trained supernet corresponding to the architecture evaluated; 2) Recalculate parameters of all batch normalization layers using a subset of the training data (we use a fixed random stratified subset of 10K samples); 3) Calculate validation and test accuracies. 

\vspace{-0.2cm}
\subsection*{Large-scale experiment}
In the large-scale experiment, we make both \emph{normal} and \emph{reduction} cells searchable. The options for \emph{normal} cells are now 5, and consist of MBConv blocks with all four possible combinations of the aforementioned expansion factors and kernels sizes ($\{3,6\} \times {\{3, 5\}}$), and the identity cell. We now further allow for the search of \emph{reduction} cells, again with 5 options. These options are the same MBConv blocks but with stride 2 and number of channels multiplication, and the same simple \emph{reduction cell} as used in our benchmarks. Recall that the positions of reduction cells are fixed (at indices 5, 10 and 15).

For this experiment we adopt the one-shot NAS approach again to speed-up the search, but include a refinement step, to better approximate the accuracy that nets would achieve if trained from scratch~\cite{Bender2018UnderstandingAS}. In particular, to evaluate an architecture, we finetune the weights obtained from the supernet for 20 epochs before calculating validation and test accuracies. Further hyperparameter settings are summarized in Table \ref{tab:training_hyperparameters}.
\newcolumntype{^}{!{\vrule width 1.5pt}}
\footnotesize
\begin{table}[H]
    \centering
    \begin{tabularx}{\textwidth}{l^p{2.6cm}|p{2.6cm}|p{2.6cm}}
         \textbf{Setup} & \textbf{Benchmarks creation} & \multicolumn{2}{l}{\textbf{Large-scale experiment}} \\ \hline
         \textbf{Hyperparameter} & \textbf{Supernet \newline training} & \textbf{Supernet \newline training} & \textbf{Single net \newline finetuning} \\ \hline
         \textbf{Optimizer} & SGD with \newline momentum 0.9, \newline weight decay $10^{-4}$ & SGD with \newline momentum 0.9, \newline weight decay $10^{-4}$ & SGD with \newline momentum 0.9, \newline weight decay $10^{-4}$ \\ \hline
         \textbf {Initial learning rate} & 0.1 & 0.1 & 0.01 \\ \hline
         \textbf{Learning rate schedule} & Cyclic cosine \newline annealing & Cyclic cosine \newline annealing & Cosine \newline annealing     \\ \hline
         \textbf{Using SWA} & \cmark & \cmark & \xmark \\ \hline
         \textbf{Number of epochs} & 370 & 500 & 20 \\ \hline
         \textbf{Batch size} & 128 & 128 & 256 \\ \hline
    \end{tabularx}
    \vspace{0.1cm}
    \caption{Training hyperparameters}
    \label{tab:training_hyperparameters}
\end{table}

\subsection*{Single-objective NAS experiments}


To provide further evidence that our proposed LS algorithm can be considered a competitive baseline in a single-objective setting as well, its performance is evaluated again on NAS-Bench-101 (CIFAR-10) and on the proposed MacroNAS-C10 and MacroNAS-C100 benchmark datasets, by optimizing for validation accuracy only.

\subsubsection*{Experimental setup.}
For these experiments LS is even simpler than before (Sec.~\ref{sec:search-algs}), as we are exclusively interested in maximizing validation accuracy and thus no longer need to sample the scalarization coefficient $\alpha$ (i.e., we fix $\alpha=1$). 
We compare again to RS, the latest single-objective version of GOMEA \cite{bosman2016expanding}, and a basic GA that uses uniform cross-over, tournament selection of size 2 upon joint parent--offspring pool, single-variable mutation with probability $p_m=1/\ell$ and population size 100. 

Similarly to Sec. \ref{sec:exp-setup}, unique architectures are evaluated only once and their accuracy is cached and re-used during the search. Besides showing the convergence of each algorithm across the NAS search spaces, we also show their performance on achieving near-optimality. We do this because finding nets with optimal validation accuracy can potentially be a considerable but pointless effort: obtaining the global optimum can be much harder than finding adequate local optima, and performance can anyhow drop when a net generalization is assessed on test data (as confirmed in Sec.~\ref{sec:generalization}). Therefore, we particularly investigate what is the effort needed for the algorithms to reach:

\begin{equation}\label{eq:acc-val-eps}
    \textit{acc}^*_{\textit{val}} - \epsilon \cdot (\textit{acc}^*_{\textit{val}} - \overline{\accval}),
\end{equation}
where $\textit{acc}^*_{\textit{val}}$ is the optimal validation accuracy, $\epsilon$ is a small number (we consider $\epsilon \in [0, 0.1]$), and $\overline{\accval}$ represents a `bottom-line' accuracy, as it is the average accuracy obtained by randomly sampling 1,000 architectures.
We particularly compute $(\textit{acc}^*_{\textit{val}} - \overline{\accval})$ to re-scale the contribution of $\epsilon$ based on the \accval{} range of the search space in exam. Chosen an $\epsilon$, the more $\overline{\accval}$ is large, i.e., the more is relatively easy to have good nets by random sampling, the more $\epsilon \cdot (\textit{acc}^*_{\textit{val}} - \overline{\accval})$ will be small, i.e., Eq.~\ref{eq:acc-val-eps} induces more proximity to the optimum.


\subsubsection*{Results.}

\begin{figure}[H]
    \centering
    \begin{subfigure}{0.32\linewidth}
        \centering
        \includegraphics[width=\linewidth]{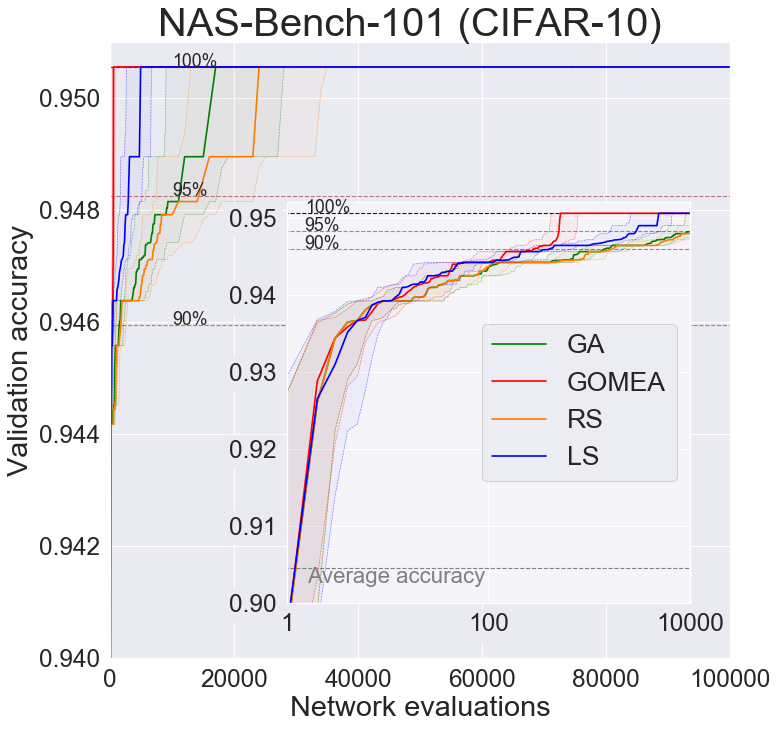}
    \end{subfigure}
    \begin{subfigure}{0.32\linewidth}
        \centering
        \includegraphics[width=\linewidth]{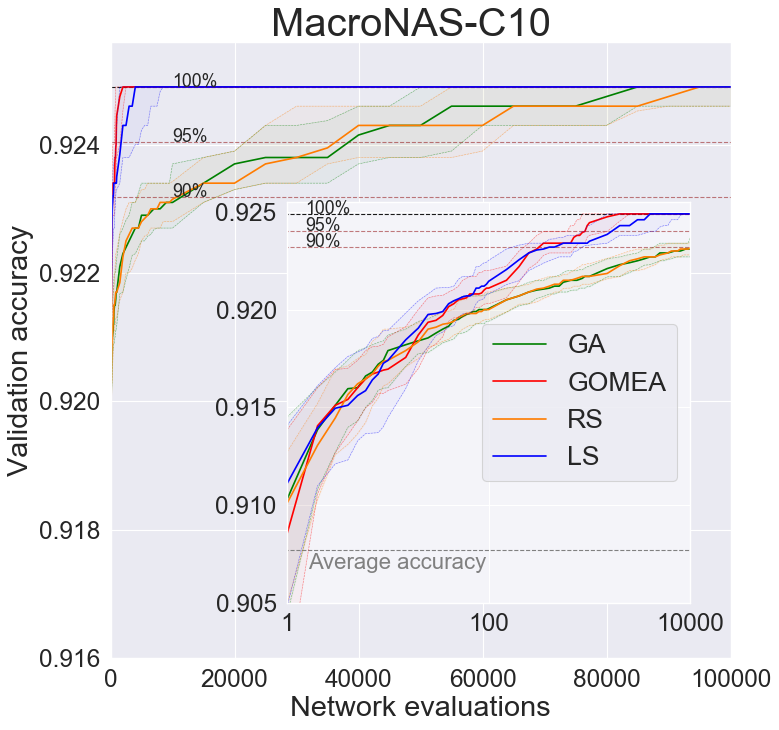}
    \end{subfigure}
    \begin{subfigure}{0.32\linewidth}
        \centering
        \includegraphics[width=\linewidth]{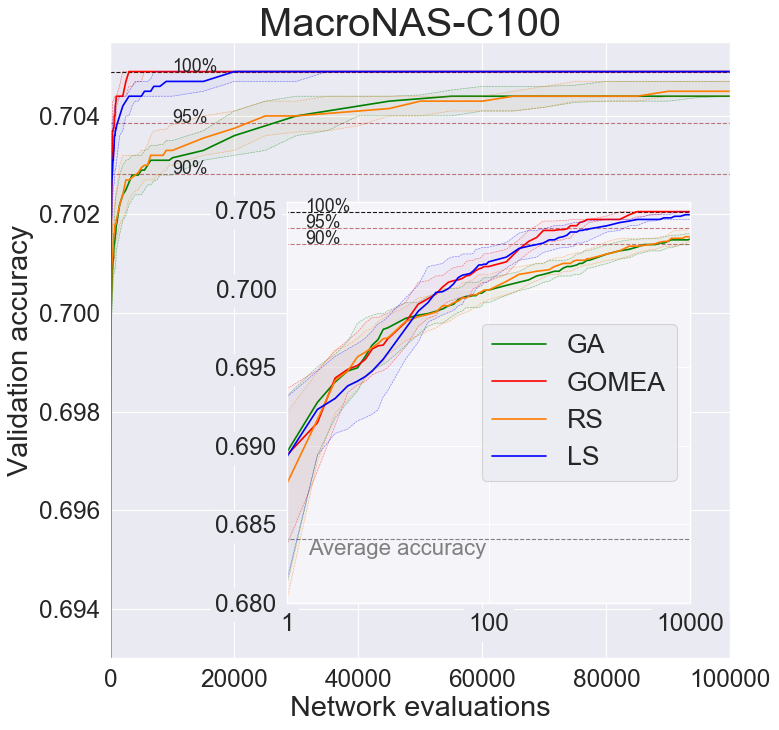}
    \end{subfigure}
    \caption{Convergence on NAS-Bench-101 (left), MacroNAS-C10 (middle) and MacroNAS-C100 (right). Medians (solid lines) and 25/75th percentiles (bands) of 100 runs are shown. Horizontal dashed lines indicate $acc_{val}^* - \epsilon \cdot (acc^*_{val} - \overline{acc_{val}})$ for $\epsilon \in \{0\%,\ 5\%,\  10\%\}$.
    }
    \label{fig:SO_convergence}
\end{figure}

Fig. \ref{fig:SO_convergence} shows convergence plots, where it can be seen that GOMEA performs best on all three search spaces, and our proposed LS algorithm is second-best. Both are consistently able to find the optimum.
The GA performs similarly to RS on MacroNAS-C10 and MacroNAS-C100, and outperforms RS on NAS-Bench-101. 
We found mutation to be essential, as without it the GA can get stuck by converging to all-identical architectures, subsequently being unable to discover a never-seen architecture.
We also found that increasing the population size by a factor of 5 still results in similar performance to RS.
Ultimately the GA is not capable of competing with GOMEA and LS.

\begin{figure}[H]
    \centering
    \begin{subfigure}{0.32\linewidth}
        \centering
        \includegraphics[width=\linewidth]{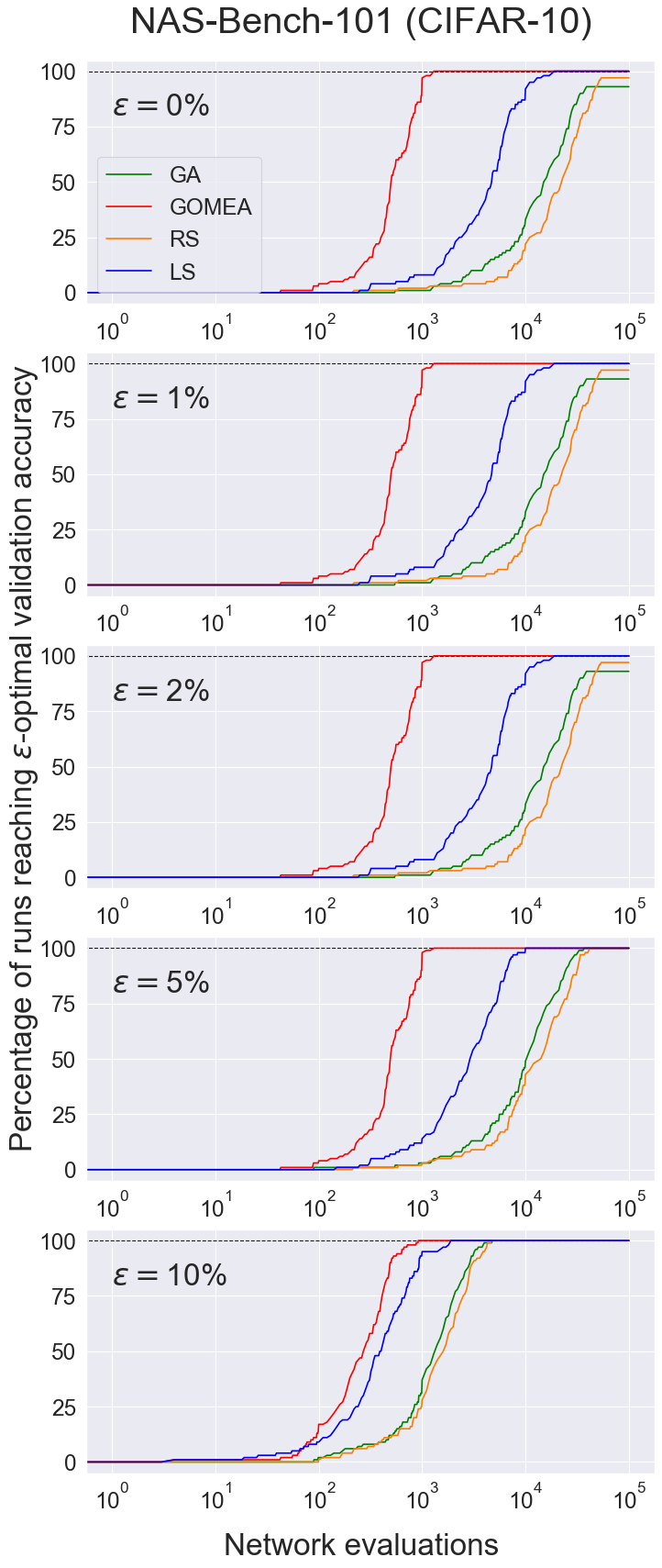}
    \end{subfigure}
    \begin{subfigure}{0.32\linewidth}
        \centering
        \includegraphics[width=\linewidth]{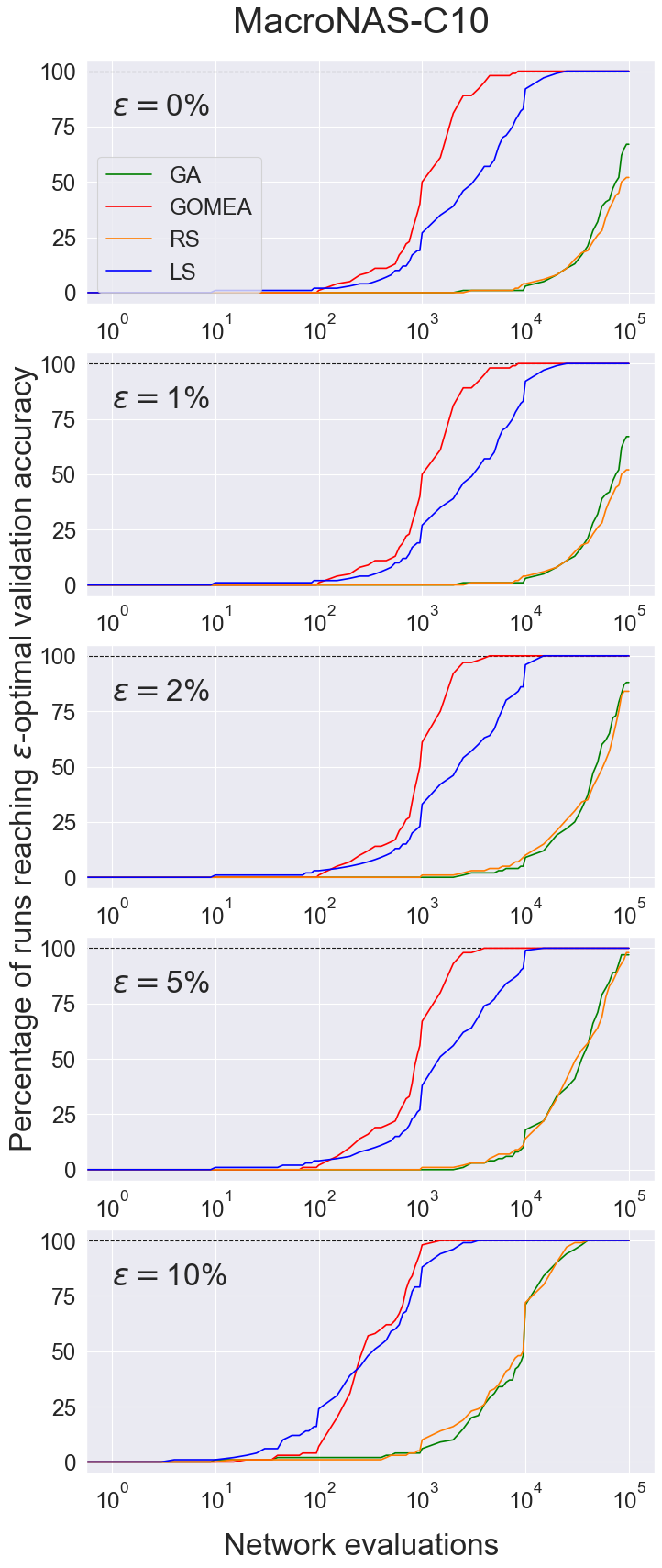}
    \end{subfigure}
    \begin{subfigure}{0.32\linewidth}
        \centering
        \includegraphics[width=\linewidth]{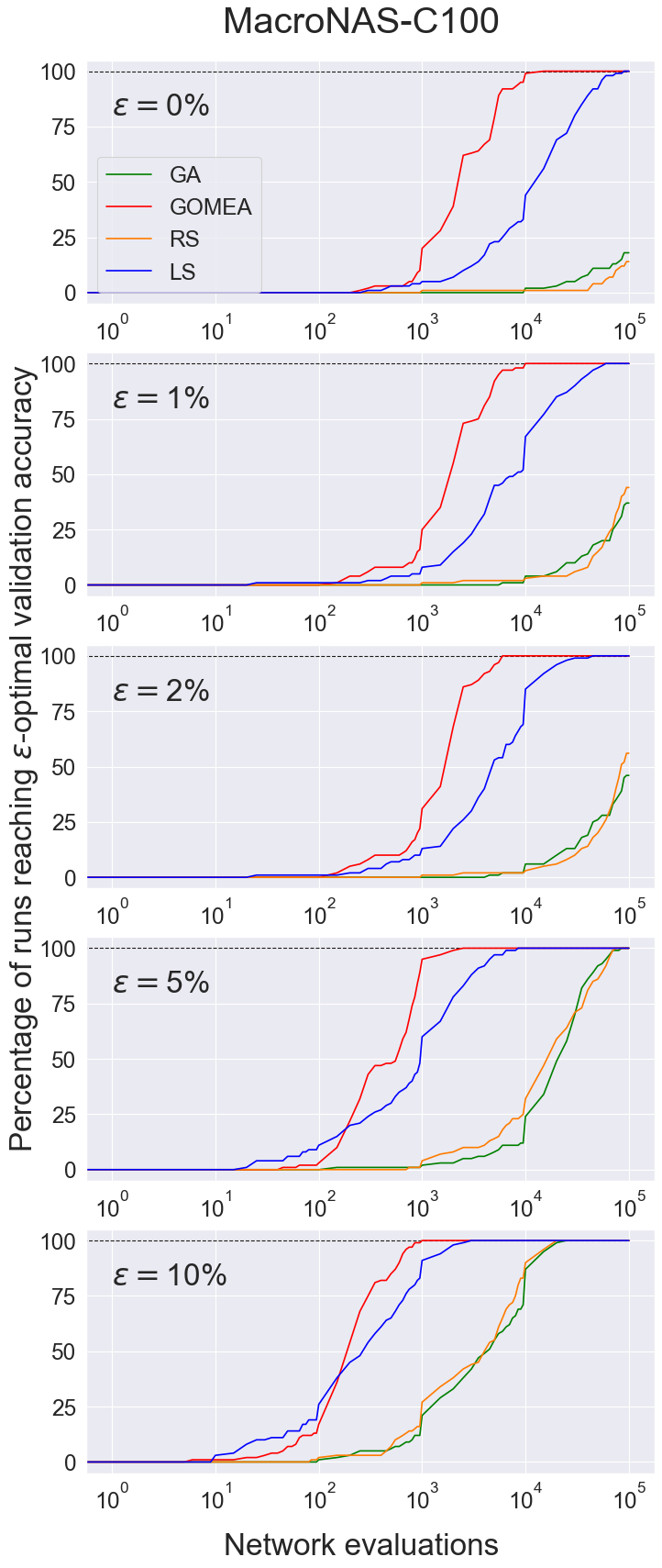}
    \end{subfigure}
    \caption{Number of runs (out of 100) reaching $acc_{val}^* - \epsilon \cdot (acc^*_{val} - \overline{acc_{val}})$ against number of network evaluations on NAS-Bench-101 (left), MacroNAS-C10 (middle) and MacroNAS-C100 (right).}
    \label{fig:SO_epsilonoptimality}
\end{figure}

In Fig. \ref{fig:SO_epsilonoptimality}, the gap in performance between GOMEA and the other algorithms becomes smaller when $\epsilon$ is larger. On these NAS search spaces, using a sophisticated search algorithm like GOMEA will result in finding the optimum (in terms of \accval) faster and more reliably. However, finding a `good' architecture instead of the optimal one is typically sufficient in NAS: the gains in \accval{} can lack substantiality (see the vertical axis of Fig.~\ref{fig:SO_convergence}), and generalization gaps (\accval{} - \acctest) can be of relatively large magnitude (Sec.~\ref{sec:generalization}). All in all, our single-objective results confirm that simpler-to-implement algorithms like LS and RS remain strong competitors. 



\clearpage

\end{document}